\documentclass[11pt]{article}

\usepackage[preprint]{acl}

\usepackage{times}
\usepackage{latexsym}

\usepackage[T1]{fontenc}

\usepackage[utf8]{inputenc}

\usepackage{microtype}

\usepackage{inconsolata}

\usepackage{graphicx}
\usepackage{enumitem}
\usepackage{caption}
\usepackage{booktabs}
\usepackage{hyperref}
\usepackage{makecell}
\usepackage{wrapfig}  
\usepackage{adjustbox}
\usepackage{tabularx}
\usepackage{mathrsfs}
\usepackage{listings} 
\usepackage{xcolor}
\usepackage{pifont}
\usepackage{alltt}  
\usepackage{subcaption}
\usepackage{url}
\usepackage{svg}
\usepackage{makecell}
\usepackage{amsmath}
\usepackage{amsfonts}
\usepackage{amssymb}
\usepackage{cleveref}
\crefname{section}{§}{§§}
\Crefname{section}{§}{§§}
\usepackage{svg}
\usepackage[normalem]{ulem} 
\usepackage{colortbl}
\usepackage{multirow}
\usepackage{algorithm}
\usepackage{algpseudocode}
\usepackage{fontawesome5}   
\usepackage[figuresright]{rotating}
\usepackage[most]{tcolorbox}
\definecolor{BoxBackground}{RGB}{240, 240, 240} 
\definecolor{BoxFrame}{RGB}{0, 0, 0} 
\definecolor{TitleBackground}{RGB}{0, 0, 0} 
\definecolor{TitleText}{RGB}{255, 255, 255} 
\tcbset{
  academicbox/.style={
    boxsep=5pt,
    left=2pt,
    right=2pt,
    bottom=0.5pt,
    boxrule=0.5pt,
    colback=BoxBackground,
    colframe=BoxFrame,
    colbacktitle=TitleBackground,
    coltitle=TitleText,
    enhanced,
    attach boxed title to top left={yshift=-0.1in,xshift=0.1in},
    boxed title style={boxrule=0pt,colframe=white},
    title={#1},
  }
}
\newtcolorbox{AcademicBox}[1][]{academicbox=#1}

\definecolor{s_doc_qa_c}{HTML}{da0d68}
\definecolor{m_doc_qa_c}{HTML}{da1d23}
\definecolor{summarization_C}{HTML}{ebb40f}
\definecolor{dialogue_c}{HTML}{187a2e}
\definecolor{synthetic_c}{HTML}{0aa3b5}
\newcommand{\Yes}{\textcolor{green!60!black}{\ding{51}}}
\newcommand{\No}{\textcolor{red!70!black}{\ding{55}}}
\title{\texttt{MemoryRewardBench}: Benchmarking Reward Models for Long-Term Memory Management in Large Language Models}

\author{
  \begin{tabular}{c} 
    Zecheng Tang$^{1,2}$, Baibei Ji$^{1,2}$, Ruoxi Sun$^{1,2}$, Haitian Wang$^{1,2}$, Wangjie You$^{1}$ \\ 
    Yijun Zhang$^{3}$, Wenpeng Zhu$^{3}$, Ji Qi$^{3}$, Juntao Li$^{1,2}$\thanks{Corresponding author.}, Min Zhang$^{1}$ 
  \end{tabular} \\
  $^{1}$Soochow University, China~~~ $^{2}$\href{https://github.com/LCM-Lab}{LCM Laboratory}  ~~~ $^{3}$China Mobile~(Suzhou), China \\
  \texttt{\{zctang, bbji\}@stu.suda.edu.cn}~~~\texttt{ \{ljt, minzhang\}@suda.edu.cn}
}

\begin{document}
\maketitle

\begin{abstract}
Existing works increasingly adopt memory-centric mechanisms to process long contexts in a segment manner~(Figure~\ref{fig:intro}), and effective memory management is one of the key capabilities that enables large language models to effectively propagate information across the entire sequence.
Therefore, leveraging reward models~(RMs) to automatically and reliably evaluate memory quality is critical.
In this work, we introduce \textbf{\texttt{MemRewardBench}}, the first benchmark to systematically study the ability of RMs to evaluate long-term memory management processes.
\texttt{MemRewardBench} covers both long-context comprehension and long-form generation tasks, featuring 10 distinct settings with different memory management patterns, with context length ranging from 8K to 128K tokens.
Evaluations on 13 cutting-edge RMs indicate a diminishing performance gap between open-source and proprietary models, with newer-generation models consistently outperforming their predecessors regardless of parameter count.
We further expose the capabilities and fundamental limitations of current RMs in evaluating LLM memory management across diverse settings.
\end{abstract}
\begin{center}
    \faGithub\ Code: \href{https://github.com/LCM-Lab/MemRewardBench}{Github link} 
    \quad
    \faDatabase\ Benchmark: \href{https://huggingface.co/datasets/LCM-Lab/MemRewardBench}{HF link}
\end{center}

\section{Introduction}
\label{sec:intro}

Large Language Models~(LLMs) have shown exceptional capabilities in comprehending contextual information~\citep{minaee2024large,xu2025towards,liu2025comprehensive}.
When tackling scenarios involving long-sequence inputs, such as long-form reasoning~\citep{bai2025longbench} or extended interactions with real-world environments~\citep{huang2025deep}, there are primarily two paradigms for processing these long sequences: (1) \emph{holistic processing} that copes with the entire sequence at once, and (2) \emph{segmented processing} that handles the sequence in chunks.
While holistic processing utilizes long context windows, segmented processing offers an efficient alternative that simultaneously supports scalable multi-turn interactions.

\begin{figure}[t]
    \centering
    \includegraphics[width=\linewidth]{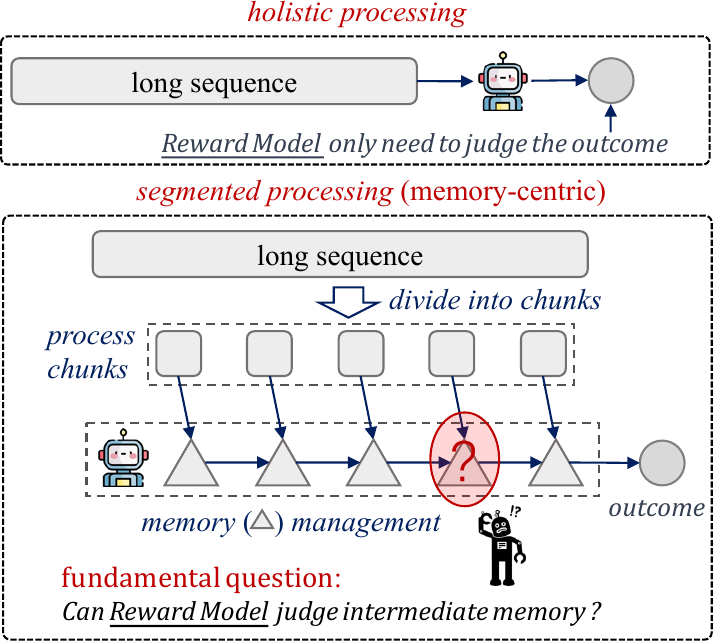}
    \caption{Illustration of holistic processing and segmented processing of long input sequence.}
    \vspace{-1em}
    \label{fig:intro}
\end{figure}

As shown in Figure~\ref{fig:intro}, segmented processing works by processing a partial segment of the context~(``chunk'' in the figure) at each step, while maintaining a fixed-size state space, i.e., memory, that summarizes historical information and integrates newly processed information~\citep{yu2025memagent,sun2025scaling,ye2025agentfold,chen2025iterresearch}.
Since the memory serves as the critical bridge between past and present information, particularly long-term information, its effective management is paramount to the model's success~\citep{xu2025mem}, necessitating the rigorous supervision of intermediate memories.
This naturally raises a fundamental question: \emph{Can we employ reward models~(RMs) to automatically evaluate intermediate memories, and what are the current boundaries of RMs in assessing memory capabilities?}

\begin{table*}[t]
\centering
\small
\resizebox{\linewidth}{!}{%
\begin{tabular}{@{}l c c c cccccc}
\toprule
\multirow{2.5}{*}{\textbf{\makecell[l]{Benchmark}}}
& \multirow{2.5}{*}{\textbf{\makecell[c]{Evaluation\\Target}}}
& \multirow{2.5}{*}{\textbf{\makecell[c]{Process\\Evaluation}}}
& \multirow{2.5}{*}{\textbf{\makecell[c]{Static vs. \\Dynamic}}}
& \multirow{2.5}{*}{\textbf{\makecell[c]{Context \\Length}}}
& \multicolumn{5}{c}{\textbf{Memory Abilities}} \\  
\cmidrule(lr){6-10}
& & & & & \textbf{DU} & \textbf{MR} & \textbf{KU} & \textbf{TR} & \textbf{GEN}\\
\midrule
LongBench \citep{longbench} & LLM & \No & Static & 0~$\sim$~64K & \No & \Yes & \No & \No & \No \\
RULER \citep{RULER} & LLM & \No & Static & 4K~$\sim$~128K & \No & \Yes & \No & \No & \No \\
\midrule
LongMemEval \citep{LongMemEval} & LLM & \Yes & Dynamic & 4K~$\sim$~115K & \Yes & \Yes & \Yes &  \Yes &  \No \\
MemoryBank \citep{MemoryBank} & LLM & \No & Static & 0~$\sim$~5K  & \Yes & \No & \No &  \Yes &  \No\\
LoCoMo \citep{LoCoMo} & LLM & \No & Dynamic & 4K~$\sim$~16K  & \Yes & \Yes & \No &  \Yes &  \Yes\\
MemBench \citep{MemBench} & LLM & \Yes & Dynamic & 0~$\sim$~100K & \Yes & \Yes & \Yes &  \Yes &  \No\\
PerLTQA \citep{PerLTQA} & LLM & \Yes & Static & 1M  & \Yes & \No & \No &  \No & \No \\
\midrule
\texttt{MemoryRewardBench}~(ours) & RM & \Yes & \makecell[c]{Static \& \\ Dynamic} & 8K~$\sim$~128K & \Yes & \Yes & \Yes & \Yes &  \Yes\\
\bottomrule
\end{tabular}%
}          
\caption{Comparison of our benchmark with existing memory benchmarks, where \textbf{DU} denotes Dialogue Understanding, \textbf{MR} denotes Multi-hop Reasoning, \textbf{KU} denotes Knowledge Update, \textbf{TR} denotes Temporal Reasoning, \textbf{GEN} denotes Generation. More details and explanations are shown in Appendix~\ref{appedix:com_benchmarks}.}
\label{tab:comparison}
\vspace{-0.5em}
\end{table*}
 
In this work, we introduce \textbf{\texttt{MemRewardBench}}, the first benchmark that is specifically designed to assess how effectively RMs judge the quality of long-term intermediate memories in LLMs. 
Unlike prior efforts that evaluate memory retention in LLMs directly (see Table~\ref{tab:benchmark_info}), we are the first to shift the focus toward benchmarking the RMs themselves, specifically their capacity to supervise and evaluate memory management.
\texttt{MemRewardBench} includes both comprehension and generation tasks, encompassing 10 diverse memory management configurations across three representative tasks: long-context reasoning, multi-turn dialogue, and long-form generation.
For each evaluation, RM is provided with the original context (ranging from 8K to 128K tokens), two candidate memory management trajectories, and their respective outcomes. 
The RM's task is to select the superior sample according to the criteria specified for each task, while also providing a justifying explanation.
To encourage RMs to \emph{prioritize memory management quality over mere outcome correctness}, we design 2 evaluation criteria to decouple the quality of the memory management process from the correctness of the outcome:
\begin{itemize}[itemsep=0pt, topsep=1pt, leftmargin=3.3em, labelsep=0.5em]
    \item[\textbf{Type 1}] \textbf{Outcome-based}: the RM should prefer a memory management trajectory that leads to a correct outcome over one that results in an incorrect outcome.
    \item[\textbf{Type 2}] \textbf{Process-based}: both memory management trajectories yield correct final outcomes, but RM should prefer the one that demonstrates more accurate, concise, and logically coherent memory updates.
\end{itemize}

We select 13 cutting-edge and widely-used LLMs as RMs, comprising 3 proprietary models and 10 open-source models.
Our results show that the performance gap between open-source and proprietary models has further narrowed.
Surprisingly, we also find that model performance does not monotonically scale with model size.
Instead, we observe a pronounced \emph{generational advantage}, whereby newer-generation models consistently outperform their predecessors regardless of parameter count, e.g., Qwen3-4B~\citep{yang2025qwen3} surpasses the substantially larger Qwen2.5-7B-Instruct~\citep{yang2024qwen2}.
Furthermore, we uncover several critical behavioral patterns in RMs, revealing both their capabilities and limitations in evaluating LLM memory management across diverse settings.



\begin{figure*}[t]
    \centering
    \includegraphics[width=\linewidth]{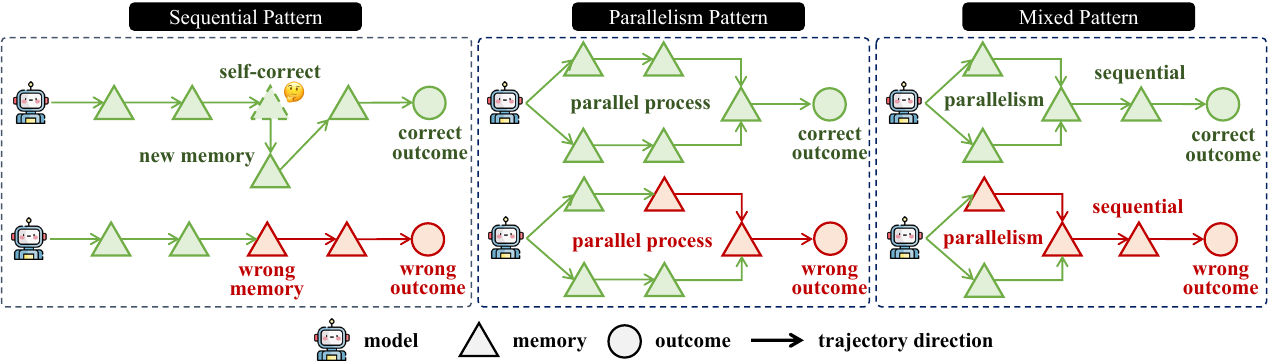}
    \vspace{-1.5em}
    \caption{Illustrations of three memory management patterns. From left to right: Sequential pattern, Parallelism pattern, and Mixed pattern. Each pattern depicts both correct and incorrect memory update trajectories. For clarity, context chunks are omitted, and only intermediate memory states are shown. More details are shown in Appendix~\ref{appdix:benchmark_construction}.}
    \label{fig:mem_pattern}
\end{figure*}  

\section{Related Work}
\label{sec:relate_work}

\subsection{Memory Management Evaluation}
\label{subsec:membench}
Existing memory management benchmarks for LLMs primarily evaluate the memories they produce and can be broadly classified into two categories. 
The first assesses memory via intermediate-state probing, which directly examines how models retain, update, and regulate memory over time~\citep{LongMemEval,MEMTRACK,MemBench}. 
This includes narrative- or domain-specific variants like StoryBench~\citep{StoryBench}, LoCoBench-Agent~\citep{LoCoBench-Agent}, and PerLTQA~\citep{PerLTQA}, which test causal coherence and sequential reasoning by adopting structured external memory to assess model robustness or evolving memory dynamics, such as MemoryBank~\citep{MemoryBank}, StuLife~\citep{stuLife}, StreamBench~\citep{StreamBench}, and Evo-Memory~\citep{Evo-Memory}.
The second category evaluates memory through \emph{final outcomes}, measuring long-term consistency in user modeling~\citep{LoCoMo,BEAM}, persona tracking~\citep{PersonaMem-v2}, or preference evolution~\citep{PrefEval}.
Extensions like Long-MT-Bench+\citep{LongMTbench} probe long-range dialogue recall, while Minerva~\citep{Minerva}, MemoryAgentBench~\citep{MemoryAgentBench}, MemoryBench~\citep{MemoryBench}, and MeetingQA~\citep{AssoMem} introduce memory-stress scenarios, stepwise fact accumulation, and large-scale context fidelity under realistic interaction settings.
Despite a wide range of existing benchmarks and evaluation efforts for LLM memory, existing approaches heavily rely on rule-based heuristics or manual annotation.
Yet, automated and scalable memory assessment paradigms based on RMs remain largely unexplored.
In this work, we address this gap by proposing \texttt{MemRewardBench}.

\subsection{Reward Model}
\label{subsec:reward_model}
Reward models~(RMs) serve as proxies for human-derived preferences, providing training signals that align language models with desired values and behaviors~\citep{bai2022training, dubois2023alpacafarm, li2023reinforcement}. 
Following the taxonomy introduced in previous work~\citep{liu2024skywork}, RM paradigms can be broadly categorized into discriminative rewards~\citep{dubois2023alpacafarm, yuan2024free, dou2025pre}, generative rewards~\citep{zheng2023judging, ligenerative2024, tang2025longrm}, and implicit rewards~\citep{rafailov2024r, xu2025distributionally}. 
Among these, generative RMs directly leverage the generalization capabilities of LLMs to generate preference judgments, thereby enabling flexible and general-purpose reinforcement learning~\citep{zhong2025comprehensive, yu2025reward}. In this work, we primarily focus on generative RMs, as this paradigm is the only one that potentially supports memory evaluation.

\section{Introduce \texttt{MemoryRewardBench}}
\label{sec:bench_construction}
Notably, different memory management strategies are adopted across tasks, and each task involves its distinct memory management patterns.
Therefore, we first identify three memory management patterns in \cref{subsec:problem_defin} and then define three task formulations in \cref{subsec:task_overview}.
Finally, we outline the data collection and benchmark construction process in \cref{subsec:construct}.

\subsection{Memory Management Pattern}
\label{subsec:problem_defin}

As shown in Figure~\ref{fig:mem_pattern}, given a model $\Phi$ and a sequence that is divided into chunks $\mathcal{C}=\{c_1, c_2, \cdots, c_n\}$, the intermediate memories $\mathcal{M}=\{m_1, m_2, \cdots, m_n\}$ are managed according to one of the following two atomic patterns:
\begin{itemize}[itemsep=0pt, topsep=0pt, leftmargin=1.5em, labelsep=0.5em]
\item \textbf{Sequential Pattern}: the memory state evolves step-by-step along the chunks, where $m_1 = \Phi(c_1), m_t=\Phi(m_{t-1}, c_t)~\mathrm{for}~ t=2,\cdots,n$, and the final outcome can be obtained from the final memory $m_n$.
\item \textbf{Parallelism Pattern}: the input context is partitioned into $k$ independent groups: $\mathcal{C}=\{\mathcal{G}_1,\cdots,\mathcal{G}_k\}$, and where each group $\mathcal{G}_j=\{c_{j,1}, \cdots,c_{j,n_j}\}$ is processed by $\Phi$ in parallel. Within each group, memory states are updated sequentially according to the \textbf{Sequential Pattern}, yielding each group's final memory state $m^{(j)}$. The final outcome is obtained by aggregating all $m^{(j)}$ through a fusion operation $g$: $o=g(m^{(1)}, \cdots, m^{(k)})$.
\end{itemize}
Notably, any memory management can be categorized as either an instance of the above two patterns, or a composition of both, i.e., the \textbf{Mixed Pattern}.

\begin{table*}[htbp]
\centering
\resizebox{\textwidth}{!}{
\begin{tabular}{l l c c c c c c c}
\toprule
\multirow{2}{*}{\bf Task Type} & \multirow{2}{*}{\bf Setting} & \multirow{2}{*}{\bf {Source}} & \multicolumn{5}{c}{\bf Length Distribution} & \multirow{2}{*}{\bf Total} \\
\cmidrule(lr){4-8}
& & & \textbf{8k} & \textbf{16k} & \textbf{32k} & \textbf{64k} & \textbf{128k} & \\
\midrule

\multirow{4}{*}{\makecell[l]{\textbf{Long-context} \\ \textbf{Reasoning}}} & Sequential-Noise & \multirow{4}{*}{\makecell[c]{BABILong~\cite{kuratov2024babilong}, \\ LongMiT~\cite{chen2025essential}}}  
                  & 101 & 44 & 43 & 36 & 31 & 255 \\
& Sequential-Drop & & 35 & 22 & 22 & 40 & 15 & 134 \\
& Mixed-Noise &   & 22 & 33 & 49 & 46 & 34 &  184 \\
& Mixed-Drop &   & 19 & 65 & 72 & 43 & 28 &  227 \\
\arrayrulecolor{black!20}\midrule

\multirow{4}{*}{\makecell[l]{\textbf{Multi-turn Dialogue} \\ \textbf{Understanding}}}  & Mem0-Out & \multirow{4}{*}{\makecell[c]{LoCoMo~\cite{maharana2024evaluating}, \\ MemoryAgentBench~\cite{hu2507evaluating}}} & 27 & 27 & 42 & 48 & 23 & 167 \\
& Mem0-Mem  & & 25 & 25 & 41 & 47 & 21 & 159   \\
& A-Mem-Out & & 42 & 42 & 48 & 50 & 47 & 229   \\
& A-Mem-Mem & & 48 & 45 & 49& 53 & 50 & 245   \\
\arrayrulecolor{black!20}\midrule

\multirow{2}{*}{\makecell[l]{\textbf{Long-form} \\ \textbf{Generation}}} & Sequential & \makecell[c]{LongEval~\cite{wu2025longeval},\\LongGenBench~\cite{wu2024longgenbench}} & 49 & 152 & 147 & 67 & 42 & 457 \\
& Parallel & LongProc~\cite{ye2025longproc} & 51 & 48 & 53 & 133 & 58  & 343 \\

\arrayrulecolor{black!20}\midrule
\rowcolor{gray!20}\textbf{Statistic} & 10 settings & - & 419 & 503 & 566 & 563 & 349  & 2400\\

\arrayrulecolor{black}\bottomrule
\end{tabular}}

\caption{Distribution and statistics of tasks in \texttt{MemoryRewardBench}, where the settings~(the ``Setting'' column) are named and defined according to the benchmark construction process described in \cref{subsec:construct}.}
\vspace{-0.5em}
\label{tab:benchmark_info}
\end{table*}

\subsection{Task Overview}
\label{subsec:task_overview}
Overall, the goal of \texttt{MemRewardBench} is to evaluate how well RMs can assess and explain the quality of an LLM’s long-term memory management according to predefined criteria. 
In this section, we outline \emph{how memory management manifests across different task settings}, and formalize \emph{the criteria that distinguish correct and incorrect memory management behaviors}.
We focus on 3 representative tasks that require memory management:
\begin{enumerate}[label=(\arabic*), itemsep=0pt, topsep=0pt, leftmargin=1.6em, labelsep=0.5em]
\item \textbf{Long-context reasoning}: the model processes a sequence of chunks $\mathcal{C}$ to extract question-relevant evidence, incrementally updating its memory, and finally produces the outcome;
\item \textbf{Multi-turn dialogue understanding}: given an extremely long conversation, e.g., hundreds of turns, the model maintains a persistent memory to record the dialogue and finally retrieves relevant dialogue turns to answer queries about a specific point in the dialogue;
\item \textbf{Long-form generation}: Given an instruction with explicit constraints, the model generates structured content over multiple steps, where intermediate generations serve as memory that must adhere to the specified constraints.
\end{enumerate}
\paragraph{RM Evaluation Criteria}
For comprehension-oriented tasks (1) and (2), the evaluation criteria for RMs are: (i) outcome-based: whether the outcome is accurate, and (ii) process-based: whether the intermediate memory is concise and relevant to the outcome. 
For the generation-oriented task (3), the key criterion is whether the intermediate memory complies with the constraints given in the instruction.
In short, even when two samples produce equally correct outcomes, one may still exhibit a superior memory management trajectory.

\subsection{Benchmark Construction}
\label{subsec:construct}
Table~\ref{tab:benchmark_info} summarizes all task types, settings, data sources, and length distributions in our benchmark. 
For each task type, we outline how we construct pairs that exhibit chosen and rejected memory management below.
Due to space limitations, we outline the core benchmark construction process below and provide further details in Appendix~\ref{appdix:benchmark_construction}.

\paragraph{Long-context Reasoning}
For a long input sequence, we employ the Sequential and the Mixed memory management patterns described in~\cref{subsec:problem_defin} to construct memory management trajectories. 
We select instances with correct final outcomes as the \emph{chosen} samples.
Then, we obtain a \emph{rejected} counterpart by introducing one of two error-inducing perturbations on the chosen sample: (1) \textsc{Noise}: injecting redundant, irrelevant information into the memory trajectory, or (2) \textsc{Drop}: dropping partial critical information from the input sequence. 
Both perturbations interfere with the memory management process, potentially compromising the final outcome. 
We regard all such memory management trajectories as rejected samples.

\begin{table*}[t]
\centering
\resizebox{\textwidth}{!}{
    \begin{tabular}{l | c c c c c | c c c c c | c c c | c }
        \toprule
    \multirow{2.5}{*}{\textbf{Models}} & \multicolumn{5}{c|}{\textbf{\emph{Long-context Reasoning}}} & \multicolumn{5}{c|}{\textbf{\emph{Multi-turn Dialogue Understanding}}}  & \multicolumn{3}{c|}{\textbf{\emph{Long-form Generation}}} & \multirow{2.5}{*}{\emph{\textbf{Avg.}}}  \\
        \cmidrule(lr){2-6} \cmidrule(lr){7-11} \cmidrule(lr){12-14} 
         & S-\textsc{Noise} & S-\textsc{Drop} & M-\textsc{Noise} & M-\textsc{Drop} & Avg. & MO & MM & AO & AM & Avg. & S & P & Avg. &\\
        \arrayrulecolor{black}\midrule
        \rowcolor{gray!20} \multicolumn{15}{c}{\textit{\faLock~ \textbf{Proprietary Models}}} \\
        \arrayrulecolor{black!20}\midrule
        
         Claude-Opus-4.5  & 47.84 & \textbf{91.05} & \textbf{52.72} & \textbf{92.51} & \textbf{68.88} & \textbf{64.07} & \textbf{45.91} & 70.61 & \textbf{84.28} & \textbf{68.25} & \textbf{89.50} & 83.97 & \textbf{87.13} & \bf 74.75 \\

        Gemini3-Pro & \textbf{54.51} & 88.81 & 48.91 & 88.99 & 68.75 & 61.08 & 45.28 & \textbf{71.43} & 82.10 & 67.13 & 75.06 & \textbf{84.26} & 79.00 & 71.63 \\

        Qwen3-Max  & 42.75 & 87.31 & 42.94 & 86.78 & 62.75 & 48.50 & 42.14 & 62.86 & 74.24 & 59.00 & 85.12 & 76.97 & 81.63 & 67.79 \\
        
        \arrayrulecolor{black}\midrule 
        \rowcolor{blue!5} \multicolumn{15}{c}{\textit{\faUnlock~\textbf{Open-source Models}}} \\
        \arrayrulecolor{black!20}\midrule    
        
    Qwen3-235A22B & 38.43 & \textbf{91.79} & 40.76 & \underline{88.99} & 62.25 & \textbf{58.08} & 40.25 & \textbf{71.18} & \underline{58.37} & \textbf{58.38} & \textbf{85.12} & 71.43 & \textbf{79.25} & \underline{66.63} \\
    GLM4.5-106A12B & \textbf{54.90} & \underline{87.31} & 49.46 & \textbf{90.31} & \textbf{69.13} & \underline{52.70} & 42.77 & \underline{59.59} & \textbf{66.38} & \underline{56.75} & \underline{79.65} & \textbf{77.55} & \underline{78.75} & \textbf{68.21} \\
    Qwen2.5-72B & 38.43 & 74.63 & 52.17 & 88.11 & 61.75 & 37.73 & 27.04 & 37.14 & 44.98 & 37.50 & 56.67 & 55.39 & 56.13 & 51.79 \\
    Llama3.3-70B & 47.06 & 70.90 & 43.48 & 76.21 & 58.50 & 52.10 & 41.51 & 52.84 & 54.69 & 51.00 & 63.46 & 62.97 & 63.25 & 57.58 \\
    Qwen3-32B & 46.67 & 82.09 & \underline{55.98} & 81.94 & 64.75 & 48.50 & \underline{43.40} & 58.08 & 56.33 & 52.63 & 71.55 & 70.85 & 71.25 & 62.88 \\
    Qwen3-14B & 49.80 & 85.82 & \textbf{56.52} & 82.38 & \underline{66.63} & 44.31 & 40.25 & 50.66 & 49.39 & 46.88 & 70.68 & 62.97 & 67.38 & 60.29 \\
    Qwen3-8B & \underline{53.73} & 68.66 & 51.09 & 72.25 & 60.88 & 28.14 & 27.04 & 48.16 & 55.90 & 42.00 & 66.08 & \underline{73.18} & 69.13 & 57.33 \\
    Llama3.1-8B & 37.65 & 53.73 & 42.94 & 61.67 & 48.38 & 41.32 & \textbf{44.03} & 35.51 & 35.81 & 38.50 & 46.17 & 43.15 & 44.88 & 43.92 \\
    Qwen2.5-7B & 28.63 & 38.81 & 39.13 & 48.90 & 38.50 & 43.71 & 32.70 & 30.20 & 23.58 & 31.63 & 47.48 & 40.23 & 44.38 & 38.17 \\
     Qwen3-4B & 53.33 & 70.90 & 46.20 & 68.72 & 59.00 & 38.32 & 30.82 & 43.27 & 49.78 & 41.63 & 56.24 & 57.14 & 56.63 & 52.42 \\
                \arrayrulecolor{black}\midrule
    \end{tabular}
}
\caption{Results on \texttt{MemoryRewardBench}, where ``S''  and  ``M'' refer to ``Sequential'' and ``Mixed'' respectively. ``MO'', ``M'', ``AO'' and ``AM'' refer to ``Mem0-\textsc{Out}'', ``Mem0-\textsc{Mem}'', ``A-Mem-\textsc{Out}'' and ``A-Mem-\textsc{Mem}'', respectively. For each metric, the best-performing result is \textbf{bolded}, and the second-best is \underline{underlined}.
}
\label{tab:main_result_full}
\end{table*}

\paragraph{Multi-turn Dialogue Understanding}
Multi-turn dialogue task exhibits strong inter-turn dependencies, necessitating that LLMs both preserve sufficient dialogue information in their memory and maintain robust temporal tracking to retrieve the most relevant memory entry~(dialogue turn) for the query.
Notably, only the Sequential memory management pattern is applicable in this task, and we adopt two dialogue memory management methods: A-Mem~\cite{xu2025mem} and Mem0~\cite{chhikara2025mem0}. 
Both methods dynamically update a summary of the dialogue history after each turn; however, A-Mem additionally annotates each summary with semantic tags (e.g., ``personal-communication'') to enable efficient memory callback.
To construct preference pairs, we select the \emph{chosen} sample based on the correctness of the final outcome. 
The \emph{rejected} sample is constructed by skipping memory updates for more than one turn of dialogue.
Notably, even when a rejected sample produces a correct final response, its intermediate memory is suboptimal due to missing or delayed updates.
For clarity, we categorize such examples into two types: samples with correct final outcomes but flawed memory management are labeled as \textsc{Mem}, while those with incorrect final outcomes are labeled as \textsc{Out}.

\paragraph{Long-form Generation}
Unlike the aforementioned comprehension tasks that provide both a question and a reference context, long-form generation supplies only an instruction with embedded constraints, requiring the model to generate content that satisfies all specified sub-constraints.
The generation process can follow either a Sequential or a Parallel memory management pattern. 
In both cases, the question is decomposed into a sequence of step-wise constraints, and the model generates content at each step to satisfy the corresponding constraint. 
The intermediate outputs are maintained as memory, where each generation is incorporated into the historical memory and conditions subsequent steps. 
After all constraints are processed, the accumulated memory states are concatenated to produce the final \emph{chosen} response.
To construct a \emph{rejected} generation, we perturb the instruction, such as dropping key constraints or injecting interference content, to cause LLMs to generate incorrect intermediate memory.

\section{Evaluation}
\label{sec:eval}

\subsection{Settings}
\label{subsec:eval_settings}
As there are currently no RMs specifically designed for evaluating memory management processes, we experiment with 13 cutting-edge LLMs as proxy RMs, including 3 proprietary models: Claude-Opus-4.5~\citep{anthropic2025claude_opus_4_5}, Gemini-3.0-Pro~\citep{deepmind2025_gemini3pro}, and Qwen3-Max~\citep{qwen2025_qwen3_max} and 10 open-source models spanning the Qwen2.5 series~\citep{yang2024qwen2}, Qwen3 series~\citep{yang2025qwen3}, and Llama3 series~\citep{dubey2024llama}, as well as GLM4.5-Air~(GLM4.5-106A12B)~\citep{zeng2025glm}. 
All RMs support a context window of at least 128K tokens.
We calculate the \emph{judgment accuracy} for each RM.
Notably, the theoretical accuracy of random guessing is 50\%. 
Yet, in practice, some RM outputs cannot be parsed, and we treat such cases as incorrect, resulting in observed accuracies falling below 50\%.
The evaluation implementation details are provided in Appendix~\ref{appdix:evaluation_settings}.

\subsection{Overall Observation}
\label{subsec:overall_observation}
We report RMs' judgment accuracy in Table~\ref{tab:main_result_full}.

\paragraph{Proprietary \emph{vs.} Open-source RMs} 
Overall, \emph{\textbf{proprietary models maintain a performance advantage}}, with Claude-Opus-4.5 achieving the highest average score of 74.75, followed closely by Gemini3-Pro at 71.63.
However, the performance gap between proprietary and open-source models has narrowed. 
GLM4.5-106A12B emerges as the strongest open-source model in our evaluation with an average score of 68.21, even outperforming the proprietary Qwen3-Max (67.79).
Specifically, we find that proprietary models remain superior in handling complex temporal dependencies and enforcing long-term constraints, as evidenced by their dominance in multi-turn dialogue and long-form generation tasks. 
Yet, open-source models close the gap in long-context reasoning reward tasks, where GLM4.5-106A12B achieves the highest score.

\begin{figure}[t]
    \centering
    \includegraphics[width=\linewidth]{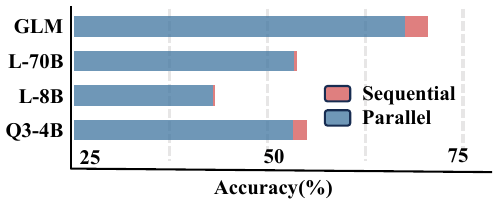}
    \caption{Performance comparison between Sequential and Parallel memory management patterns on the long-context reasoning and long-form generation tasks. }
    \label{fig:sp_comparison}
\end{figure}

\paragraph{Open-source RMs Analysis}
The performance of open-source RMs reveals a pronounced \emph{\textbf{decoupling between parameter count and practical capability}}, underscoring the impact of more efficient training data curation and increasingly effective post-training strategies in the latest generation of models.
This trend is particularly evident in the Qwen3 series, where Qwen3-32B~(62.88) not only outstrips much larger models such as Llama3.3-70B~(57.58) but also marginally exceeds its own larger variant, Qwen3-235A22B~(66.63).
From another perspective, the newer-generation models significantly outperform their predecessors, e.g., Qwen3-8B (57.33) achieves a substantial performance gain over the previous-generation Qwen2.5-7B (38.17). 
This improvement is likely attributable to advances in context-scaling training and post-training strategies adopted in newer models, which may foster more robust reasoning processes that align more closely with the judgment-and-explanation paradigm required by RM evaluation.

\paragraph{Cross-Task Capability Characterization}
The comparison across task categories reveals differences in task difficulty and model strengths. 
Multi-turn dialogue is the most challenging task, consistently yielding lower RM scores due to the need for RMs to accurately perceive conversational state transitions in order to assess the correctness of intermediate memory.
Long-form generation is moderately difficult, as it requires RMs to assess whether memory updating process exhibits sustained adherence to global constraints throughout the generation process.
On both tasks, proprietary RMs maintain a performance lead, and cutting-edge open-source RMs demonstrate competitive results.
In contrast, long-context reasoning appears to be the most tractable task, consistently achieving the highest overall scores across RMs.
This suggests that retrieving and reasoning over static information has become a relatively mature capability for current LLMs, and \emph{\textbf{effective management of dynamic memory and long-range constraints}} remains the key factor distinguishing top-performing RMs.

\section{Ablation Study}
\label{sec:analysis}
In this section, we analyze RM behavior from four perspectives:
(1) LLM memory management patterns~(\cref{subsec:effect_mem_manage_pattern});
(2) RM evaluation criteria: distinguishing between outcome-based and process-based judgments as well as examining RM robustness to global constraints~(\cref{subsec:effect_outcome_process_reward});
(3) RM sensitivity to memory management trajectory length~(\cref{subsec:effect_memory_management_trajectory}); and
(4) the impact of memory-enhancement strategies on RM performance~(\cref{subsec:impact_mem_model_strategy}).
We hope the above experiments provide insights for applying and improving RMs for evaluating LLM memory.

\paragraph{RM Selection and Notation}
We primarily select the following RMs for analysis and introduce shorthand notations for clarity: GLM-4.5-106A12B (GLM), Llama-3.3-70B-Instruct (L-70B), Llama-3.1-8B-Instruct (L-8B), and various sizes of Qwen3 (Q3), e.g., Q3-4B denotes Qwen3-4B.

\begin{figure*}[t]
  \centering
  \begin{subfigure}[t]{0.49\textwidth}
    \centering
     \includegraphics[width=\linewidth]{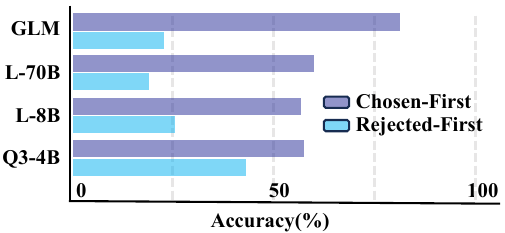}
    \caption{Both chosen and rejected samples have correct outcome, but rejected one has a redundant memory management trajectory.}
    \label{fig:position_bias_self_correct}
  \end{subfigure}
  \hfill
  \begin{subfigure}[t]{0.49\textwidth}
    \centering
    \includegraphics[width=\linewidth]{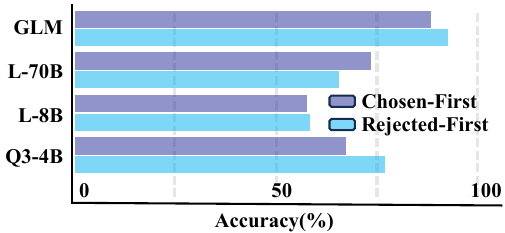}
    \caption{Chosen sample has correct outcome and rejected sample has wrong outcome.}
    \label{fig:position_bias_mask_info}
  \end{subfigure}
  \caption{Comparison between process-based and outcome-based reward criteria. Chosen-First indicates that the chosen sample is presented before the rejected sample in the input context fed to the RM, and vice versa.} 
  \label{fig:outcome_process_compare}
\end{figure*}

\begin{figure}[t]
    \centering
    \includegraphics[width=\linewidth]{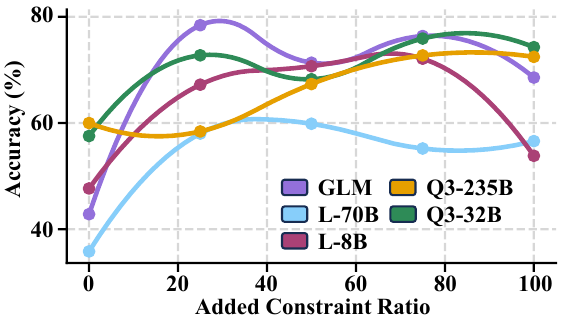}
    \caption{Performance trends of RMs with increasing constraint density in long-form generation instructions.}
    \label{fig:longgen_constraint}
\end{figure}

\begin{figure*}[t]
\centering
\begin{subfigure}[t]{\textwidth}
\centering
\includegraphics[width=\linewidth]{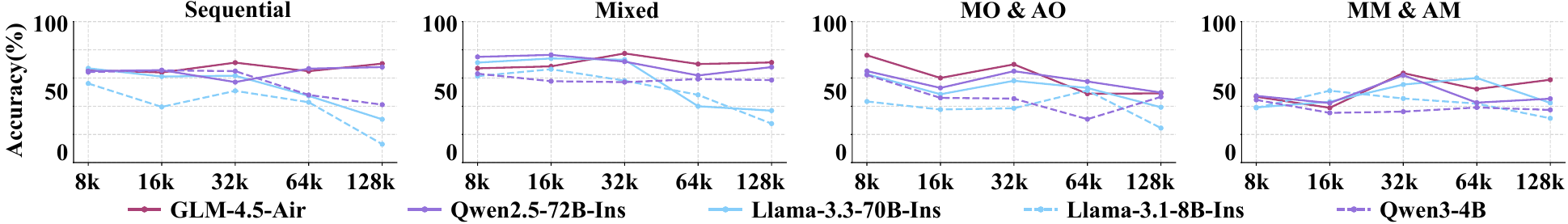}
\caption{RM performance~(Accuracy) on \texttt{MemRewardBench} as context length increases.}
\label{fig:length_induced_bias_acc}
\end{subfigure}
\begin{subfigure}[t]{\textwidth}
\centering
\includegraphics[width=\linewidth]{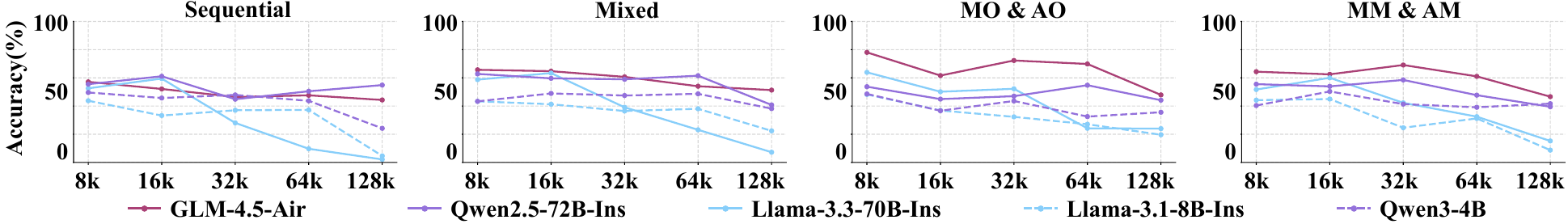}
\caption{RM consistency~(Accuracy) on \texttt{MemRewardBench} as context length increases.}
\label{fig:length_induced_bias_cons}
\end{subfigure}
\caption{Trends in RM performance and consistency with respect to memory management trajectory length. The 1st and 2nd columns correspond to the \emph{Long-context Reasoning} task, the 3rd column to the \emph{Multi-turn Dialogue Understanding} task (average score), and the 4th column to the \emph{Long-form Generation} task (average score).}
\label{fig:length}
\end{figure*}

\subsection{Effect of Memory Management Patterns}
\label{subsec:effect_mem_manage_pattern}
We compare the performance of RMs under Sequential and Parallelism memory management pattern s on long-context reasoning and long-form generation tasks. 
As shown in Figure~\ref{fig:sp_comparison}, results show that RMs achieve significantly higher accuracy under the Sequential pattern. 
This suggests that \emph{current RMs exhibit a stronger preference for progressive, step-by-step reasoning processes}, which aligns more closely with the causal structures commonly present in models' training data and language modeling~\citep{jiao2025think}. 
In contrast, RMs struggle to effectively evaluate outputs generated through parallel processing and subsequent merging, highlighting a notable limitation and a promising direction for future improvement. 
More results are provided in Appendix~\ref{appdix:memory_management_patterns}.

\subsection{Effect of RM Evaluation Criteria}
\label{subsec:effect_outcome_process_reward}

\paragraph{Outcome- \emph{vs.} Process-based Criterion}
Evaluating memory management requires RMs to assess not only outcomes but also the quality of intermediate memory states. 
To investigate whether RMs prioritize outcome-based or process-based signals when evaluating memory management, we compare their behavior under outcome-based and process-based reward modeling paradigms.
We adopt RM \textbf{evaluation consistency} as the primary metric. 
Specifically, we swap the positions of the chosen and rejected samples in the RM's input context and evaluate the RM twice: once with the original ordering and once with the reversed ordering, to assess whether the RM's preference remains stable under this positional perturbation.
We perform the above evaluation under two settings: (a) \emph{process-based}, where both outcomes are correct but differ in the quality of their memory management trajectories; and (b) \emph{outcome-based}, where only one sample produces a correct outcome.
As shown in Figure~\ref{fig:outcome_process_compare}, RMs exhibit inconsistency in the process-based setting, displaying a positional bias that favors samples appearing earlier in the input context.
In contrast, under the outcome-based setting, RMs show robust and consistent preferences aligned with the ground truth\footnote{In our main experiments, the chosen and rejected samples are randomly ordered, which mitigates the impact of positional bias and ensures the overall evaluation results remain reliable.}.

\paragraph{Adherence to Global Constraint}
RMs are expected to rigorously evaluate outcomes based on all constraints specified in the instruction. 
To assess RMs' global constraint coherence, we evaluate performance under progressively increasing constraint densities, ranging from topic-only prompts to fully specified, multi-constraint instructions.
Implementation details are provided in Appendix~\ref{appdix:long_form_gen_detail_ablation}.
As shown in Figure~\ref{fig:longgen_constraint}, RM performance initially improves as more instructional detail is incorporated, peaking at approximately 25\% constraint density. 
At this level, RMs benefit from sufficient explicit criteria to ground their judgments in concrete instruction-following signals. 
However, further increases in constraint density do not yield continued gains; instead, performance plateaus or even declines. 
This suggests that current RMs are only partially capable of leveraging dense, multi-faceted constraints to assess memory fidelity.

\subsection{Effect of Memory Management Trajectory Length}
\label{subsec:effect_memory_management_trajectory}
We first plot the trend of RM performance as context length increases in Figure~\ref{fig:length_induced_bias_acc}.
Within a context length of up to 64K tokens, most RMs can maintain accuracy above 50\%.
Then, we evaluate RM consistency across different context length intervals using the position-swapping protocol described in \cref{subsec:effect_outcome_process_reward}. 
Our results show that only GLM-4.5-Air and Qwen2.5-72B-Instruct maintain stable performance, achieving above 50\% accuracy in most context lengths.
In contrast, other models fail to maintain accuracy above 50\%, exhibiting pronounced inconsistency once the context length exceeds 32K tokens.
While this degradation can be attributed to limited parameter size, e.g., Qwen3-4B, models from the Llama family, particularly Llama-3.3-70B-Instruct, exhibit severe performance collapse at 64K and 128K context length, despite their substantially larger parameter count. 
We provide a detailed case analysis of such abnormal behavior and report comprehensive model performance across all tasks in Appendix~\ref{appdix:abnormal_case}.

\begin{figure}[t]
        \includegraphics[width=\linewidth]{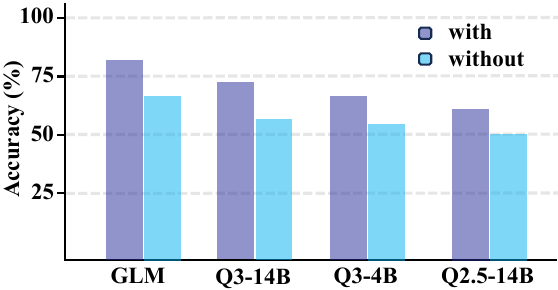}
        \caption{Comparison on multi-turn dialogue understanding task with and without auxiliary signals.}
        \label{fig:tag_impact_on_models}
\end{figure}

\subsection{Effect of Memory Augmentation Strategy}
\label{subsec:impact_mem_model_strategy}
Finally, we evaluate RMs under augmented memory management settings, aligning with the core focus of recent work that seeks to improve performance by introducing additional constraints or enhancements to memory mechanisms~\citep{lestable2025}. 
We conduct experiments in the challenging multi-turn dialogue understanding task to assess RMs' performance.
One way to enhance memory management in multi-turn dialogue understanding is to annotate each memory update with a semantic tag, e.g., personal-communication, that characterizes the contextual nature of the dialogue segment~\citep{xu2025mem}. 
To investigate the impact of such auxiliary signals on RMs' evaluation accuracy, we compare RM performance under two settings: memory updates with explicit tags and memory updates without tags.
We show the above data structure in Appendix~\ref{appdix:multiturn_dialogue_under_tag}. 
As shown in Figure~\ref{fig:tag_impact_on_models}, incorporating auxiliary signals consistently improves the accuracy of RMs in evaluating memory management quality. 
Combined with the analysis in \cref{subsec:effect_memory_management_trajectory}, semantic tags provide RMs with concise, high-level summaries of dialogue context, thereby enabling more reliable judgment without requiring the model to parse potentially redundant or verbose memory update trajectories.

\section{Conclusion} 
Automatically evaluating the long-term memory management process of LLMs is essential. 
In this work, we introduce \texttt{MemRewardBench}, the first benchmark designed to systematically assess how effectively current RMs evaluate LLM long-term memory management.
Our evaluation shows that open-source RMs have largely closed the gap with proprietary models on long-context reasoning, but still lag behind on tasks with long-range dependencies, such as multi-turn dialogue understanding and memory-intensive long-form generation. 
Our analysis further highlights both the strengths and fundamental limitations of current RMs in evaluating LLM memory management.
We hope \texttt{MemRewardBench} provides a valuable benchmark and offers practical guidance for improving reward modeling and advancing memory-centric LLMs.



\bibliography{main}

\appendix

\clearpage

\section{Comparison between \texttt{LongRewardBench} and Existing Memory Benchmarks}
\label{appedix:com_benchmarks}

Table~\ref{tab:comparison} provides an overall comparison between \texttt{LongRewardBench} and representative existing memory benchmarks in terms of evaluation targets, evaluation paradigms, task settings, and the dimensions of memory ability they cover.
Below, we will clarify key terminology and provide a more detailed explanation of the comparison criteria.

\paragraph{Evaluation Target} refers to the primary object of assessment. Most existing memory benchmarks take LLMs as the direct evaluation target and measure memory capability based on the correctness or consistency of model outputs. In contrast, \texttt{LongRewardBench} focuses on RMs, evaluating how effectively RMs judge the quality of long-term intermediate memories in LLMs.

\paragraph{Process Evaluation} distinguishes whether a benchmark explicitly evaluates intermediate states during task execution. Benchmarks marked with ``\No'' focus solely on outcome's correctness and do not assess intermediate reasoning or memory update processes, whereas those marked with ``\Yes'' evaluate intermediate steps. \texttt{LongRewardBench} emphasizes the assessment of long-term intermediate memory trajectories and therefore supports process-level evaluation.

\paragraph{Static \emph{vs.} Dynamic} characterizes whether task-relevant information changes over time. Static settings assume a fixed context or memory state, requiring the model to reason over unchanging information. In contrast, Dynamic settings involve continual information updates during interaction or generation, where the model must maintain, revise, or overwrite existing memory. \texttt{LongRewardBench} incorporates both static and dynamic scenarios: Multi-turn Dialogue Understanding is inherently dynamic due to the continuous integration of temporal signals and new interactions, whereas Long-context Reasoning and Long-form Generation are treated as static. This design better reflects realistic memory-management requirements in practical deployments.

\paragraph{Context Length} specifies the range of context lengths involved in each benchmark, reflecting the extent to which long-term memory capability is evaluated. Compared with benchmarks constrained to shorter context windows, \texttt{LongRewardBench} covers a broader range of long contexts, enabling a more systematic analysis of RMs' long-horizon memory evaluation capabilities.

In the \textbf{Memory Abilities} column, we map the task types covered by different benchmarks to five core memory capabilities:
\begin{itemize}[itemsep=0pt, topsep=0pt, leftmargin=1.5em, labelsep=0.5em]
\item \textbf{DU (Dialogue Understanding)}, which evaluates a model’s ability to comprehend multi-turn dialogue histories and maintain consistency across interactions;
\item \textbf{MR (Multi-hop Reasoning)}, which assesses the model’s capacity to perform reasoning and information integration across multiple inference steps;
\item \textbf{KU (Knowledge Update)}, which focuses on the model’s ability to update existing knowledge or internal memory upon the introduction of new information;
\item \textbf{TR (Temporal Reasoning)}, which examines the model’s ability to model and reason about temporal order, event sequencing, and evolutionary processes;
\item \textbf{GEN (Generation)}, which evaluates the model’s ability to maintain content coherence and satisfy multiple constraints in long-form or multi-stage generation tasks.
\end{itemize}
Existing memory benchmarks typically cover only a subset of the aforementioned memory abilities, whereas \texttt{LongRewardBench} integrates multiple task formulations to achieve systematic coverage of diverse memory capabilities within a unified evaluation framework. Specifically, DU and GEN are directly assessed through the Multi-turn Dialogue Understanding and Long-form Generation tasks, respectively. In the Long-context Reasoning task, the incorporation of multi-hop reasoning data enables the evaluation of both MR and TR, while the use of the MemAgent~\citep{yu2025memagent} paradigm in Long-context Reasoning further equips the task with the ability to assess KU. Through these designs, \texttt{LongRewardBench} provides a more comprehensive and fine-grained characterization of RMs' memory evaluation capabilities compared to other existing memory benchmarks.

\begin{table*}[htbp]
    \centering

\resizebox{\textwidth}{!}{
\begin{tabular}{l l l >{\raggedright\arraybackslash}m{2.8cm} c l}
\toprule
\bf Task & \bf \makecell[l]{LLM Memory \\ Management} & \bf Type & \bf {\makecell[l]{Preference \\ Construction}} & \bf Dataset & \bf Construction Description \\
\midrule
\multirow{6}{*}{\makecell[l]{Long-context \\ Reasoning}} & \multirow{6}{*}{\makecell[l]{MemAgent \\ \citep{yu2507memagent}}} & \multirow{3}{*}{Sequential} & \multirow{2}{*}{Self-Correct} & BABILong & \multirow{3}{*}{\makecell[l]{Process context sequentially, Self-Correct denotes \\ real-time correction of memory. Drop-Info denotes \\ masking key information.}} \\
& &    &    & LongMIT & \\
\arrayrulecolor{gray!20}\cmidrule(lr){4-5} 
&  &  &  Drop-Info  & LongMIT &  \\
\arrayrulecolor{gray!20}\cmidrule(lr){3-6} 

& &  \multirow{3}{*}{Parallel}  & \multirow{2}{*}{Self-Correct} & BABILong & \multirow{3}{*}{\makecell[l]{Process context in parallel, then aggregate memories \\ of each line. }} \\
& &   &     & LongMIT &  \\
\arrayrulecolor{gray!20}\cmidrule(lr){4-5} 
&  &      &  Drop-Info   & LongMIT &  \\

\arrayrulecolor{black!20}\midrule

\multirow{4}{*}{\makecell[l]{Multi-turn \\ Dialogue \\ Understanding}} & \multirow{4}{*}{\makecell[l]{Mem0\\ \citep{xu2025mem} \\ A-Mem\\ \citep{chhikara2025mem0}}} & \multirow{4}{*}{Sequential} & \multirow{2}{*}{OUT} & LoCoMo & \multirow{4}{*}{ \makecell[l]{Classify the samples based on whether the answer of \\ the rejected is correct or not. OUT  is 0 as \\ easy. MEM is classified as hard. }} \\
& &                             &    & MemoryAgentBench & \\
\arrayrulecolor{black!20}\cmidrule(lr){4-5} 

& & & \multirow{2}{*}{MEM}     & LoCoMo & \\
& &                             &    & MemoryAgentBench & \\

\arrayrulecolor{black!20}\midrule

\multirow{5}{*}{\makecell[l]{Long-form \\ Generation}} & \multirow{5}{*}{-} & \multirow{3}{*}{Sequential} & Direct-Generate & LongProc & Plan routes based on the given  mode of transportation. \\

\cmidrule(lr){4-6} 
& &  & \multirow{2}{*}{Prompt-Modify} & LongEval & Generate a complete article based on the given outline. \\ 
& &  & & LongGenBench &  Generate long text based on given constraints. \\ 
\cmidrule(lr){3-6} 
& & \multirow{2}{*}{Parallel} & \multirow{2}{*}{Prompt-Modify}  & LongEval & Generate a complete article based on the given outline. \\ 
& &  &  & LongGenBench &  Generate long text based on given constraints. \\
\arrayrulecolor{black}\bottomrule
\end{tabular}

}
\caption{Overview of construction details of \texttt{MemRewardBench}. Due to space limitation, we provide partial citations here: BABILong~\citep{kuratov2024babilong}, LongMIT~\citep{chen2025essential}, LoCoMo~\citep{LoCoMo}, MemoryAgentBench~\citep{MemoryAgentBench}, LongProc~\citep{ye2025longproc}, LongEval~\citep{wu2025longeval}, and LongGenBench~\citep{wu2024longgenbench}.}
\label{appdix:tab:benchmark_source}
\end{table*}
\begin{figure*}[t] 
\begin{AcademicBox}[A\_Mem and Mem0  Case]
\footnotesize

\textbf{\textit{[A\_Mem system]}} \\
\textbf{Question:} \\
What is James' favorite game? \\

\textbf{A\_Mem\_Chosen:} \\
\textbf{[Excerpt from a larger conversation memory pool]} \\
\textbf{[conversation\_72]} \\
\textbf{Context:} The main topic is about playing the game Apex Legends with a team, emphasizing the intensity and sharing a screenshot of the gameplay. \\
\textbf{Keywords:} ['Apex Legends', 'game', 'team', 'screenshot', 'intense'] \\
\textbf{Tags:} ['gaming', 'social media', 'entertainment', 'team gameplay'] \\

\textbf{A\_Mem\_Rejected:} \\
\textbf{[Excerpt from a larger conversation memory pool]} \\
\textbf{[conversation\_72]} \\
\textbf{Context:} The main topic is about a player sharing their experience and excitement playing the multiplayer video game Apex Legends with their team, and inviting others to view a screenshot of their gameplay. \\
\textbf{Keywords:} ['game', 'team', 'screenshot', 'playing'] \\
\textbf{Tags:} ['Video Games', 'Gaming', 'Team Play'] \\

\textbf{Answer:} \\
James's favorite game is Apex Legends.

\vspace{4pt}
\hrule
\vspace{6pt}

\textbf{\textit{[Mem0 system]}} \\
\textbf{Question:} \\
What is James interested in? \\

\textbf{Mem0\_Chosen:} \\
\textbf{[Representative memory excerpt]} \\
\textbf{'id':} '72eae632-3310-4ec0-93c5-97d535408d2f', \\
\textbf{'memory':} 'Jame enjoys reading books.', \\
\textbf{'hash':} 'a65072fab28c8d0ef7aff048a3af4ae1', \\
\textbf{'metadata':} None, \\
\textbf{'created\_at':} '2025-12-01T20:37:46.562661-08:00' \\

\textbf{Mem0\_Rejected:} \\
\textbf{[Representative memory excerpt]} \\
\textbf{'id':} '367e4e8b-8cd9-4266-a2e4-e33dc8c09b18', \\
\textbf{'memory':} "[5:35 pm on 22 March, 2023] Speaker Jame says: ``I’ve recently developed a strong interest in reading. I enjoy immersing myself in books, especially how reading allows me to explore different ideas, stories, and perspectives in a focused and reflective way.''", \\
\textbf{'hash':} '8544b92aab390170f4f19d309e6d192a', \\
\textbf{'metadata':} None, \\
\textbf{'created\_at':} '2025-12-04T07:04:52.138025-08:00' \\

\textbf{Answer:} \\
Reading books.

\vspace{6pt}
\hrule
\vspace{6pt}

\end{AcademicBox}

\caption{Illustration of chosen and rejected memory excerpts in a\_mem and mem0 systems under different question types.}
\label{fig:memory_selection_case}
\end{figure*}

\begin{figure*}[t]
\begin{AcademicBox}[LongProc Case]
\footnotesize
\textbf{\textit{[sequential scheme]}} \\
\textbf{Question:}\\
\textit{[TASK]}\\
In a completely hypothetical world, there are a number of cities. Each city has a one-way connection to only one other city via a specific transit method (bus, train, plane, or ferry). Your task is to provide a route from a city to another city. You should follow the specific instruction provided later and output the route following the format provided in the instruction…\\
\textit{[PROBLEM]}\\
In a hypothetical world, there are a number of cities. Each city has a one-way connection to only one other city via a specific transit method. The details of the cities are as follows: …Garland is a lively city. You can travel from Garland to Royston by ferry. Ivanovo is a lively city. You can travel from Ivanovo to Frome by plane. Frome is a lively city. You can travel from Frome to Swift Current by bus. Ostrava is a lively city. You can travel from Ostrava to Frome by train. Rennes is a lively city. You can travel from Rennes to Royston by bus. \textbf{Royston is a lively city. You can travel from Royston to Redhill by plane}…\\
Now find the route from Warsaw to Burnley based on the information above. Some reminders: - All connections are one-way. You can solve the problem by iteratively finding the next city to travel to until you reach the destination city. - Follow the specific format for the route output. Mark the route with <Route> and </Route> tags. - Describe in detail how you determine the route step by step (at least 550 sentences)."\\
\textbf{Chosen:} \\
From Warsaw, take a plane to Lichfield. From Lichfield, take a ferry to Bern. From Bern, take a bus to Porto. From Porto, take a train to Rennes. From Rennes, take a bus to Royston. \textbf{From Royston, take a plane to Redhill}. From Redhill, take a plane to Hitchin…From Argenteuil, take a train to Kingston. From Kingston, take a plane to Vienna. From Vienna, take a train to Burnle.\\
\textbf{Rejected:} \\
From Warsaw, take a plane to Lichfield. From Lichfield, take a ferry to Bern. From Bern, take a bus to Porto. From Porto, take a train to Rennes. From Rennes, take a bus to Royston. \textbf{From Royston, take a plane to Frome}. From Frome, take a bus to Swift Current…From Argenteuil, take a train to Kingston. From Kingston, take a plane to Vienna. From Vienna, take a train to Burnley.\\

\vspace{-5pt} \hrule \vspace{4pt}
\end{AcademicBox}

\caption{LongProc Case. }
\label{appdix:longproc_case}
\end{figure*}
\begin{figure*}[t]
\begin{AcademicBox}[LongEval Case]
\footnotesize
\textbf{\textit{[Sequential scheme]}} \\
\textbf{Question:}\\
Title: Building Conversational AI Agents By Integrating Reasoning, Speaking \& Acting With LLMs\\
please generate a blog based on the given bullet point: …\textbf{Employing AI Agents for engaging conversations with customers}…\\
\textbf{Chosen:} \\
Based on the title Building Conversational AI Agents By Integrating Reasoning, Speaking \& Acting With LLMs, \textbf{the previously generated text}, and the given bullet points …\textbf{Employing AI Agents for engaging conversations with customers… , I have composed the following section.} …These advanced AI agents are designed not just to perform routine operations but to \textbf{engage in meaningful and dynamic interactions with customers}…\\
\textbf{Rejected:} \\
Based on the title Building Conversational AI Agents By Integrating Reasoning, Speaking \& Acting With LLMs ,\textbf{the previously generated text}, and the given bullet points …\textbf{Employing AI Agents for automated decision-making without human intervention}…,  I have composed the following section. …these agents can understand context, make informed decisions, and \textbf{execute actions with minimal human oversight}…\\

\hrule \vspace{4pt}

\textbf{\textit{[Parallel scheme]}} \\
\textbf{Question:}\\
Title: Microsoft’s Magnetic-One: The Dawn of Truly Autonomous AI\\
please generate a blog based on the given bullet point: … \textbf{The system is capable of performing various tasks such as booking flights and writing code}…\\
\textbf{Chosen:} \\
Based on the title Microsoft’s Magnetic-One: The Dawn of Truly Autonomous AI and the given bullet points …\textbf{The system is capable of performing various tasks such as booking flights and writing code}…, I have composed the following section … from \textbf{booking flights and writing code} to potentially even making a mean cup of coffee …\\
\textbf{Rejected:} \\
Based on the title Microsoft’s Magnetic-One: The Dawn of Truly Autonomous AI and the given bullet points …\textbf{The system is capable of performing various tasks, but it primarily focuses on entertainment and gaming}…, I have composed the following section …While its primary focus is currently on \textbf{entertainment and gaming}, the potential applications are vast …\\

\vspace{-5pt} \hrule \vspace{4pt}
\end{AcademicBox}

\caption{LongEval Case. }
\label{appdix:longeval_case}
\end{figure*}
\begin{figure*}[t]
\begin{AcademicBox}[LongGenBench Case]
\footnotesize
\textbf{\textit{[sequential scheme]}} \\
\textbf{Question:}\\
As Chef Roy Hee, a world-renowned chef at a globally renowned restaurant, you are tasked with designing an entire year's menu for 2018. Your menu should be varied and innovative, adhering to the following guidelines:…\textbf{2) Feature 'Autumn Harvest Fest' from Week 15 to Week 16 with dishes like: Roasted Squash and Root Vegetables}…\\
\textbf{Chosen:} \\
Based on the events: Autumn Harvest Fest featuring Roasted Squash and Root Vegetables, \textbf{the previously generated text}, I have composed the following section. Menu Week 15 (April 9th - April 15th) Welcome to the \textbf{**Autumn Harvest Fest**}! …2. Main Courses: 2.1) \textbf{**Roasted Squash and Root Vegetables**}: A medley of butternut squash, parsnips, and carrots, roasted to perfection with thyme and rosemary. …\\
\textbf{Rejected:} \\
Based on the events: Autumn Harvest Fest featuring Roasted Squash and Root Vegetables, \textbf{the previously generated text}, I have composed the following section. Menu Week 15 (April 9th - April 15th) This week’s menu is a tribute to the bold and hearty flavors of the Middle East, where spices, grains, and legumes come together in harmony…2. Main Courses: 2.1) Lamb Kofta with Sumac and Mint: Spiced lamb meatballs with a bright, herbaceous garnish. 2.2) Grilled Salmon with Za’atar and Sumac: A zesty, aromatic fish dish with a tangy, smoky flavor.\\

\hrule \vspace{4pt}

\textbf{\textit{[Parallel scheme]}} \\
\textbf{Question:}\\
Noah is a historian with a passion for chronicling his vibrant life through weekly diary entries. Captures:…\textbf{3) Participate in a language exchange meetup every 5 weeks on weekends, starting from week 13}….\\
\textbf{Chosen:} \\
Based on the events: Participate in a language exchange meetup, I have composed the following section. Week 13 (March 25th - March 31st):…\textbf{This week also marked the start of my language exchange meetups, which take place every five weeks on weekends.} I attended the first session on the 24th, and it was both exciting and slightly intimidating to engage in conversation with people from all over the world….\\
\textbf{Rejected:} \\
Based on the events: Participate in a language exchange meetup, I have composed the following section. Week 13 (March 25th - March 31st): The air is finally beginning to feel like spring, and the days are lengthening in a most welcome way. Temperatures have warmed to a pleasant 12°C by the end of the week, coaxing the first hints of green to peek through the remaining patches of snow in the park.\\

\vspace{-5pt} \hrule \vspace{4pt}
\end{AcademicBox}

\caption{LongGenBench Case. }
\label{appdix:longgenbench_case}
\end{figure*}

\begin{figure*}[t]
\begin{AcademicBox}[S-Drop Case]
\footnotesize
\textbf{Question:}\\
Here is a long context but only a few clues are useful, while others are interferences, you should answer the question following the context after carefully reading it. 

[Context]
Bassel El-Gharbawy ...
Abdelouahed Idrissi Chorfi
Abdelouahed Idrissi Chorfi (born 1 January 1969) is a Moroccan judoka.  He competed at the 1996 Olympics as an extra-lightweight, coming seventeenth.
Achievements
References

Category:1969 births
Category:Living people
Category:Moroccan male judoka
Category:Judoka at the 1996 Summer Olympics ...
Majemite Omagbaluwaje (born 10 June 1969) is a Nigerian judoka. He competed at three Olympic Games.

[Question]Which athlete has a more diverse set of achievements, Bassel El-Gharbawy or Majemite Omagbaluwaje, considering their participation in various games?\\

\textbf{Chosen:} \\
Step1:

<problem> 
Which athlete has a more diverse set of achievements, Bassel El-Gharbawy or Majemite Omagbaluwaje, considering their participation in various games?
</problem>

<memory>
No previous memory
</memory>

<section>
Bassel El-Gharbawy
Bassel El-Gharbawy (born 10 April 1977) is an Egyptian judoka. ...
</section>

<wrong-updated-memories>
No wrong updated memory
</wrong-updated-memories>

Updated memory:
<updated-memory>
Bassel El-Gharbawy (born 10 April 1977) is an Egyptian judoka. ...
</updated-memory>

Explanation:
I have updated the memory by retaining the correct information from the wrong updated memory and the section, ...

Step2: ...
Step3: ...
Step4: ...
Step5: ...

Updated memory:
<problem> 
Which athlete has a more diverse set of achievements, Bassel El-Gharbawy or Majemite Omagbaluwaje, considering their participation in various games?
</problem>

<memory>
... I will continue to gather more information to make a more informed decision.

Comparing Maj \\

\textbf{Rejected:} \\
Step1:

<problem> 
Which athlete has a more diverse set of achievements, Bassel El-Gharbawy or Majemite Omagbaluwaje, considering their participation in various games?
</problem>

<memory>
No previous memory
</memory>

<section>
Abdelouahed Idrissi Chorfi

Abdelouahed Idrissi Chorfi (born 1 January 1969) is a Moroccan judoka.  He competed at the 1996 Olympics as an extra-lightweight, coming seventeenth. ...

</section>

<wrong-updated-memories>
No wrong updated memory
</wrong-updated-memories>

Updated memory:
<updated-memory> 
...
No information is available about Bassel El-Gharbawy or Majemite Omagbaluwaje, so it's not possible to compare their achievements.

Step2:

<problem> 
Which athlete has a more diverse set of achievements, Bassel El-Gharbawy or Majemite Omagbaluwaje, considering their participation in various games?
</problem>

<memory>
<updated-memory> ...
</memory>
<section>
None
</section>

<wrong-updated-memories>
No wrong updated memory
</wrong-updated-memories>

Updated memory:
Since we have no information about Bassel El-Gharbawy and Majemite Omagbaluwaje, and the previous wrong answer was based on an incorrect inference, we cannot make a comparison between their achievements.

It is not possible to compare their achievements as there is no information available about Bassel El-Gharbawy and Majemite Omagbaluwaje. \\

\vspace{-5pt} \hrule \vspace{4pt}
\end{AcademicBox}
\vspace{-1em}
\caption{S-Drop Case. }
\label{appdix:s_drop_case}
\end{figure*}
\begin{figure*}[t]
\begin{AcademicBox}[P-Noise Case]
\footnotesize
\textbf{Question:}\\
Here is a long context but only a few clues are useful, while others are interferences, you should answer the question following the context after carefully reading it. [Context]
 Sandra travelled to the bedroom. ...  Sandra grabbed the apple there. ... Sandra gave the apple to John. John journeyed to the bathroom. ... John discarded the milk. ... Sandra went back to the bathroom. John put down the apple there. ... . Daniel went to the bathroom...
[Question]How many objects is John carrying? \\
\textbf{Chosen:} \\
The entire agent workflow consists of parallel processing components and a summarization processing component.
Each parallel processing component is as follows:
[Part 1]
Step 1: ...
Step 2: ...

<wrong-updated-memories>
No wrong updated memory
</wrong-updated-memories>

Updated memory:
<updated-memory>
There is no information about John in the provided section. ...
From the given section, we can see that the narrative is focused on the little girl, Mary, Daniel, and old Hans. ...
Upon re-examining the section, I noticed that there is no mention of John whatsoever. ...
The scene is indeed taking place in a rural setting, with the sound of a creek and old Hans' axe in the background. The background sounds do not provide any information about John's presence or actions. 
Therefore, my updated memory is that there is still no information about John, ...

[Part 2]
Step 1: ...
Step 2: ...
Updated memory:
<updated-memory>
There is a house with a courtyard and stables situated on the fourth side of the quadrangle. The stables are not being used much, as the horses are grazing in an adjoining paddock. John is seen grabbing something in the stables, specifically the milk. ...
</updated-memory>

[Part 3]
Step 1: ...
Step 2: ...
Updated memory:
<updated-memory>
John was carrying an apple initially. He then placed it down in the bathroom. There is no information about John carrying any other objects.
</updated-memory>

After processing in parallel, the summarization processing component is as follows:
There are 3 processed components (Memories):
[Part 1]
 <updated-memory>
There is no information about John in the provided section. However, it does not necessarily mean he is not carrying any objects. We would need more context or information about John to determine the number of objects he is carrying. ...
[Part 2]
 <updated-memory>
There is a house with a courtyard and stables situated on the fourth side of the quadrangle. The stables are not being used much, as the horses are grazing in an adjoining paddock. ...
</updated-memory>
[Part 3]
 <updated-memory>
John was carrying an apple initially. He then placed it down in the bathroom. There is no information about John carrying any other objects.
</updated-memory>
The summarization of them is as follows: There is no information about John in the provided section. ... The text also mentions that Sandra grabbed an apple and gave it to John. ...

Based on the memory, the process of obtaining the answer is as follows:
<problem> 
How many objects is John carrying? 
</problem>
<memory>
<updated-memory>
John was carrying an apple initially. He then placed it down in the bathroom. There is no information about John carrying any other objects.
</updated-memory>
</memory>

Updated memory:
Based on the given memory, John was initially carrying an apple. He then placed it down in the bathroom, which means he is no longer carrying it. 
Therefore, John is carrying 0 objects. \\

\textbf{Rejected:} \\
The entire agent workflow consists of parallel processing components and a summarization processing component.
Each parallel processing component is as follows:
[Part 1]
Step 1: ...
Step 2: ...

<wrong-updated-memories>
\textbf{ There is no information about John in the provided section. ...
From the given section, we can see that the narrative is focused on the little girl, Mary, Daniel, and old Hans. ...
Upon re-examining the section, I noticed that there is no mention of John whatsoever. ...
The scene is indeed taking place in a rural setting, with the sound of a creek and old Hans' axe in the background. The background sounds do not provide any information about John's presence or actions. 
Therefore, my updated memory is that there is still no information about John, ...}
</wrong-updated-memories>

Updated memory:
<updated-memory>
There is no information about John in the provided section. ...
From the given section, we can see that the narrative is focused on the little girl, Mary, Daniel, and old Hans. ...
Upon re-examining the section, I noticed that there is no mention of John whatsoever. ...
The scene is indeed taking place in a rural setting, with the sound of a creek and old Hans' axe in the background. The background sounds do not provide any information about John's presence or actions. 
Therefore, my updated memory is that there is still no information about John, ...

[Part 2]
Step 1: ...
Step 2: ...
Updated memory:
<updated-memory>
There is a house with a courtyard and stables situated on the fourth side of the quadrangle. The stables are not being used much, as the horses are grazing in an adjoining paddock. John is seen grabbing something in the stables, specifically the milk. ...
</updated-memory>

[Part 3]
Step 1: ...
Step 2: ...
Updated memory:
<updated-memory>
John was carrying an apple initially. He then placed it down in the bathroom. There is no information about John carrying any other objects.
</updated-memory>

After processing in parallel, the summarization processing component is as follows:
There are 3 processed components (Memories):
[Part 1]
 <updated-memory>
There is no information about John in the provided section. However, it does not necessarily mean he is not carrying any objects. We would need more context or information about John to determine the number of objects he is carrying. ...
[Part 2]
 <updated-memory>
There is a house with a courtyard and stables situated on the fourth side of the quadrangle. The stables are not being used much, as the horses are grazing in an adjoining paddock. ...
</updated-memory>
[Part 3]
 <updated-memory>
John was carrying an apple initially. He then placed it down in the bathroom. There is no information about John carrying any other objects.
</updated-memory>
The summarization of them is as follows: There is no information about John in the provided section. ... The text also mentions that Sandra grabbed an apple and gave it to John. ...

Based on the memory, the process of obtaining the answer is as follows:
<problem> 
How many objects is John carrying? 
</problem>
<memory>
<updated-memory>
John was carrying an apple initially. He then placed it down in the bathroom. There is no information about John carrying any other objects.
</updated-memory>
</memory>

Updated memory:
Based on the given memory, John was initially carrying an apple. He then placed it down in the bathroom, which means he is no longer carrying it. 
Therefore, John is carrying 0 objects. \\

\vspace{-5pt} \hrule \vspace{4pt}
\end{AcademicBox}
\vspace{-1em}
\caption{M-Noise Case. The bold parts are redundant noise. }
\label{appdix:p_noise_case}
\end{figure*}
\begin{figure*}[t]
\begin{AcademicBox}[M-Drop Case]
\footnotesize
\textbf{Question:}\\
Here is a long context but only a few clues are useful, while others are interferences, you should answer the question following the context after carefully reading it. [Context]
A man was seriously wounded in a Long Beach shooting, authorities said Monday.

The shooting was reported at 10:40 p.m. Sunday at Atlantic Avenue and Burnett Street, Long Beach police Lt. James Richardson said.

Arriving officers found a man shot in the upper body, Richardson said.

Paramedics took the man to a hospital where he was in serious condition, the lieutenant said. ...

[Question]What was the condition of the man found by the arriving officers, where was he taken, and what are the details of the incident? \\

\textbf{Chosen:} \\
The entire agent workflow consists of parallel processing components and a summarization processing component.
Each parallel processing component is as follows:
[Part 1]
Step 1: ...
Step 2: ...

[Part 2]
Step 1: ...
Step 2: ...

After processing in parallel, the summarization processing component is as follows:
There are 2 processed components (Memories):
[Part 1]
A shooting incident occurred at 10:45 p.m. Tuesday night in the 2700 block of East Seventh Street in Long Beach. Multiple suspects were at large after the shooting. The incident was described as a "mass-casualty incident" and was initially reported as a house party shooting. However, it's unclear whether the shooting occurred during the party or before it. Three men were kil...
[Part 2]
The incident involved gunfire at a Long Beach home in Southern California, killing at least three people and wounding several others, possibly at a Halloween party that was being held at the house. ...
Based on the memory, the process of obtaining the answer is as follows:

<problem> 
What was the condition of the man found by the arriving officers, where was he taken, and what are the details of the incident?
</problem>

<memory>
Updated memory:

The incident involved gunfire at a Long Beach home in Southern California, killing at least three people and wounding several others, possibly at a Halloween party that was being held at the house. The exact number of people shot is not immediately available, with reports ranging from 7 to 10 people. ...
</memory>

Updated memory:
Based on the provided memory and problem, I will answer the question.

The condition of the man found by the arriving officers is not specified. They were taken to a nearby yard outside the Long Beach home. The incident involved a shooting at a house party in Long Beach, Southern California, where three people were killed and nine others were hospitalized. \\

\textbf{Rejected:} \\
The entire agent workflow consists of parallel processing components and a summarization processing component.
Each parallel processing component is as follows:
[Part 1]
Step 1: ...

[Part 2]
Step 1: ...
Step 2: ...

After processing in parallel, the summarization processing component is as follows:
There are 2 processed components (Memories):
[Part 1]
The updated memory will be based on the provided section, as there is no previous memory to retain.

Three men were killed in a shooting incident at a Halloween party in the 2700 block of East Seventh Street in Long Beach on Tuesday night. 
The incident occurred around 10:45 p.m. 
Multiple suspects were at large following the shooting. ...

[Part 2]
 The man found by the arriving officers was among the nine people injured in the shooting. 

The condition of the man is not immediately available, however, it is mentioned that some of the wounded, including the man, suffered life-threatening gunshot wounds. ...

<problem> 
What was the condition of the man found by the arriving officers, where was he taken, and what are the details of the incident?
</problem>

<memory>
The man found by the arriving officers was among the nine people injured in the shooting. 

The condition of the man is not immediately available, however, it is mentioned that some of the wounded, including the man, suffered life-threatening gunshot wounds. ...
</memory>

Updated memory:
Based on the provided memory and problem, I will answer the question.

The condition of the man found by the arriving officers is life-threatening as he suffered a gunshot wound. The location where he was taken is not specified in the article. The details of the incident include a shooting at a Long Beach home in Southern California, resulting in at least three deaths and nine injuries, with the shooter firing indiscriminately into a crowd of 25 to 30 people from an alley behind the residence. \\

\vspace{-5pt} \hrule \vspace{4pt}
\end{AcademicBox}
\vspace{-1em}
\caption{M-Drop Case. }
\label{appdix:p_drop_case}
\end{figure*}

\section{Benchmark Construction}
\label{appdix:benchmark_construction}
Overview of \texttt{MemRewardBench} construction statistics is shown in Table~\ref{appdix:tab:benchmark_source}.
We provide the construction details below.

\subsection{Long-context Reasoning} 
\paragraph{Prototype Description}
We first define the key information as the content to which the model must attend in order to correctly answer the question, and define the contexts that include key information as key contexts. As described in \cref{subsec:construct}, we introduce two error-inducing perturbations: \textsc{Noise} and \textsc{Drop}. We apply these two methods based on \texttt{MemAgent}~\cite{yu2507memagent}, which processes a long context by chunking it into equal segments, iteratively updating its memory after each segment to condense key information into a fixed-size buffer for response generation. We construct ``Sequential'' pattern data based on the MemAgent's sequential processing mechanism, and the ``Mixed'' pattern is built upon the ``Sequential'' pattern.

\paragraph{Dataset Description}
We build up our Long-Context Reasoning task on Babilong~\cite{kuratov2024babilong} and LongMiT~\cite{ chen2025essential}, for the convenience of extracting clue documents and background documents. For LongMiT, we take the clue documents as the key contexts. For babilong, we take the needles as the key contexts.

\paragraph{\textsc{Noise} Perturbation}
We define noise as any \emph{redundant information} introduced by the MemAgent during the memory updating process. 
In this work, we consider two types of redundancy: (1) incorrect memory updates and (2) repeated memory updates.
To induce incorrect memory updates, we employ weaker models (Llama-3.1-8B-Instruct/Qwen2.5-7B-Instruct) as the MemAgent engine. 
After each memory update, we adopt an LLM-as-judge paradigm, using a stronger model(Qwen3-235A22B) to assess the quality of the updated memory. 
If the update is judged correct, it is passed to the next chunk; otherwise, the engine model is instructed to revise the memory according to the feedback. 
This correction process is repeated whenever erroneous updates persist. Samples that either never fail or fail to be corrected after 10 attempts are discarded. 
Upon completion of the above process, we cache the full inference trajectory, extract the complete trajectory as the chosen sample, and remove the incorrect updates from it to construct the corresponding rejected sample.
Then, to induce redundancy caused by repetition, we reuse the previously discarded never-failed samples as chosen samples. For each such trajectory, we randomly insert an additional memory segment extracted from the same trajectory, thereby introducing redundant repetition to form the rejected sample. For more details, see the case shown in Figure~\ref{appdix:s_noise_case}.

\begin{figure*}[t]
\begin{AcademicBox}[S-Noise Case]
\footnotesize
\textbf{Question:}\\
Here is a long context but only a few clues are useful, while others are interferences, you should answer the question following the context after carefully reading it. [Context]
Early math knowledge is as important as early literacy for children's subsequent achievement. ... Now a new longitudinal study has found that young children whose mothers supported them during play, specifically in their labeling of object quantities, had better math achievement at ages 4-½ and 5 years. ...

[Question]What did a new longitudinal study find about the impact of mothers supporting their children's labeling of object quantities on math achievement?
\\
\textbf{Chosen:} \\
Step1:...
Step2:
<problem> 
What did a new longitudinal study find about the impact of mothers supporting their children's labeling of object quantities on math achievement?
</problem>
<memory>
<updated-memory>
Early math knowledge is crucial for children's subsequent achievement, predicting later school success better than early reading skills and even income in adulthood. ...
Mothers' support during play, specifically in labeling object quantities, was found to have a positive impact on math achievement at ages 4-½ and 5 years. ...
To assess mothers' support of their children's math skills, researchers examined how moms supported and guided their 3-year-olds' learning as they played with a toy cash register and blocks. ...
</memory>
<section>
While many studies have been conducted on infants' and preschoolers' math competencies, few have evaluated how toddlers' ...
Geary and his team including Alex Moore, a postdoctoral fellow, and Kristy vanMarle, an assistant professor in the ...
"We measured participants' math skills at the beginning of preschool and again at the end of preschool," Geary said. ...
Geary and his team will continue to follow their participants through first grade in this four-year longitudinal study where ....
</section>
<wrong-updated-memories>
No wrong updated memory
</wrong-updated-memories>

Updated memory:
<updated-memory>
Early math knowledge is crucial for children's subsequent achievement, predicting later school success better than early reading ...
Mothers' support during play, specifically in labeling object quantities, was found to have a positive impact on math achievement at ages 4-½ and 5 years. ...
To assess mothers' support of their children's math skills, researchers examined how moms supported and guided their 3-year-olds' learning as they played ...

Step3:...

Updated memory:
<problem>
What did a new longitudinal study find about the impact of mothers supporting their children's labeling of object quantities on math achievement?
</problem>
<memory>
... (same as provided)
</memory>
<wrong-answer>
No wrong answer
</wrong-answer>

Since the problem is about the impact of mothers supporting their children's labeling of object quantities on math achievement, and the memory provides information about a study conducted by researchers at Boston College, which found that mothers' support during play, specifically in labeling object quantities, had a positive impact on math achievement at ages 4-½ and 5 years.

Therefore, the answer is:
Mothers' support during play, specifically in labeling object quantities, has a positive impact on math achievement.\\

\textbf{Rejected:} \\
Step1: ...
Step2:
<problem> 
What did a new longitudinal study find about the impact of mothers supporting their children's labeling of object quantities on math achievement?
</problem>
<memory>
<updated-memory>
Early math knowledge is crucial for children's subsequent achievement, predicting later school success better than early reading skills and even income in adulthood. ...
Mothers' support during play, specifically in labeling object quantities, was found to have a positive impact on math achievement at ages 4-½ and 5 years. ...
To assess mothers' support of their children's math skills, researchers examined how moms supported and guided their 3-year-olds' learning as they played with a toy cash register and blocks. ...
</memory>
<section>
While many studies have been conducted on infants' and preschoolers' math competencies, few have evaluated how toddlers' ...
Geary and his team including Alex Moore, a postdoctoral fellow, and Kristy vanMarle, an assistant professor in the ...
"We measured participants' math skills at the beginning of preschool and again at the end of preschool," Geary said. ...
Geary and his team will continue to follow their participants through first grade in this four-year longitudinal study where ....
</section>
<wrong-updated-memories>
\textbf{<updated-memory>
Early math knowledge is crucial for children's subsequent achievement, predicting later school success better than early reading skills and even income in adulthood. ...
Mothers' support during play, specifically in labeling object quantities, was found to have a positive impact on math achievement at ages 4-½ and 5 years. ...
To assess mothers' support of their children's math skills, researchers examined how moms supported and guided their 3-year-olds' learning as they played with a toy cash register and blocks. ...}
</wrong-updated-memories>

Updated memory:
<updated-memory>
Early math knowledge is crucial for children's subsequent achievement, predicting later school success better than early reading skills and even income in adulthood. ...
Mothers' support during play, specifically in labeling object quantities, was found to have a positive impact on math achievement at ages 4-½ and 5 years. ...
To assess mothers' support of their children's math skills, researchers examined how moms supported and guided their 3-year-olds' learning as they played with a toy cash register and blocks. ...

Step3: ...

Updated memory:
<problem>
What did a new longitudinal study find about the impact of mothers supporting their children's labeling of object quantities on math achievement?
</problem>
<memory>
... (same as provided)
</memory>
<wrong-answer>
No wrong answer
</wrong-answer>

Since the problem is about the impact of mothers supporting their children's labeling of object quantities on math achievement, and the memory provides information about a study conducted by researchers at Boston College, which found that mothers' support during play, specifically in labeling object quantities, had a positive impact on math achievement at ages 4-½ and 5 years.

Therefore, the answer is:
Mothers' support during play, specifically in labeling object quantities, has a positive impact on math achievement.\\

\vspace{-5pt} \hrule \vspace{4pt}
\end{AcademicBox}
\vspace{-1em}
\caption{S-Noise Case. The bold parts are redundant noise. }
\label{appdix:s_noise_case}
\end{figure*}

\paragraph{\textsc{Drop} Perturbation}
Another approach to inducing errors is dropping key information from the context.
Specifically, we remove all key information while retaining the full background context, ensuring that the resulting memories lack crucial evidence. 
This setup simulates scenarios in which the agent ignores essential clues; the corresponding trajectory is used to construct the rejected sample. 
In contrast, we apply the standard MemAgent process to the complete context to obtain the chosen sample.
Throughout this procedure, we ensure that key information is always contained within a single, complete chunk rather than split across adjacent chunks, guaranteeing that its removal leads to an incorrect response.
We further filter out samples whose chosen trajectories fail to produce correct answers, thereby avoiding cases where both chosen and rejected samples omit critical information and exhibit insufficient preference separation. For more details, see the case shown in Figure~\ref{appdix:s_drop_case}.

\paragraph{Implementation Details of Mixed Pattern}
Following the same procedure as before, we first divide the long context into segments of equal length. 
As illustrated on the right group of Figure~\ref{fig:mem_pattern}, each segment is initially processed independently under a parallel pattern. For the parallel part of the mixed pattern, let $p$ be the number of chunks divided, referring to the parallel size, we set $p \in \{2, 3\}$. We then introduce an aggregation mechanism to integrate the memories from all segments. 
After aggregation, the agent continues to update its memory sequentially to generate the final answer, adhering to the sequential pattern throughout this stage. 
Since this process combines both parallel and sequential memory management patterns, we categorize the resulting preference pairs as Mixed-Noise or Mixed-Drop, respectively. For more details, see the case shown in Figure~\ref{appdix:p_noise_case}, ~\ref{appdix:p_drop_case}.

\subsection{Multi-turn Dialogue Understanding} 
As described in \cref{subsec:impact_mem_model_strategy}, we evaluate RM performance on the multi-turn dialogue understanding task with and without auxiliary signals. We focus on two representative memory systems, A-Mem~\cite{xu2025mem} and Mem0~\cite{chhikara2025mem0}, both of which are designed for long-dialogue scenarios and operate by dynamically storing and iteratively updating memories across conversation turns. However, the two systems differ substantially in how memory is organized and maintained.

A-Mem assigns semantic tags to different segments of a conversation at each round to summarize their content. When a new conversation round begins, the system retrieves and updates the top-$k$ most relevant memories based on the current dialogue context. In contrast, Mem0 does not employ a tagging mechanism. Instead, it maintains a global memory summary, into which new information is directly incorporated at each update step(As illustrated by the examples in Fig.~\ref{fig:memory_selection_case}).

\paragraph{Pair-data Construction}  
The conversational data used in the experiment comes from the Locomo~\cite{LoCoMo} dataset and the Conflict\_Resolution dataset from MemoryAgentBench~\cite{MemoryAgentBench}, which focuses on relationship information and key statements within the conversations.

For sample generation, we produce both positive and negative samples from the conversation data. Positive samples are created by processing all rounds of the conversation and ensuring that the final memory chain leads to a correct answer. Only memory chains that result in accurate answers are retained as positive samples.

Negative samples are generated by manipulating the frequency of memory update triggers, intentionally leaving some updates incomplete (e.g., skipping updates in certain rounds). This results in memory chains that are less complete and less well-organized compared to the corresponding positive samples. These incomplete memory chains are further classified into two categories:

\textbf{(Mem) Negative Samples}: If the memory chain retains the key information required to answer the question but has flaws in memory management (such as reduced efficiency in information retrieval), and the system still outputs the correct answer, the memory chain is marked as {Mem}, indicating "correct result but with memory management defects."

\textbf{(Out) Negative Samples}: If the memory chain is so incomplete that key information is lost or difficult to retrieve, leading to an incorrect answer, it is marked as {Out}, indicating "incorrect result due to failure in key information retrieval."

\subsection{Long-form Generation}
\paragraph{Prototype Description}
As described in Section~\ref{subsec:construct}, Long-form Generation is modeled as a constraint-driven task, where the model iteratively generates content based on a series of progressive constraints. 
The preference settings for long text generation tasks are based on the pathtraversal subset of LongProc~\cite{ye2025longproc}, as well as the LongGenBench~\cite{wu2024longgenbench} and LongEval~\cite{wu2025longeval} datasets.

\textbf{(1) Longproc}: 
The pathtraversal subtask of LongProc provides the model with all known routes along with specified start and end points, and requires it to generate the correct sequence of steps leading from the source to the destination. 
This task emphasizes procedural reasoning, as each intermediate state is explicitly dependent on the preceding generation step, rendering the process inherently sequential. 
Accordingly, parallel generation is not applicable to LongProc, and only the sequential generation scheme is employed for this dataset.

\textbf{(2) LongGenBench and LongEval}: 
In contrast, LongGenBench takes a set of clearly specified constraints as input, including detailed content requirements and overall structural conditions.
Similarly, LongEval uses instructions containing paragraph-level constraints to define the generation requirements for different parts of the article.
For both datasets, the model needs to generate a long text output that satisfies all the given constraints.
Due to the decomposable input and segmented output characteristics of these two datasets, we applied both sequential and parallel generation schemes.

\paragraph{Benchmark Construction Based on LongProc}
For the path traversal subset of LongProc datasets, only employ a sequential generation scheme. A representative example is presented in Figure~\ref{appdix:longproc_case}.

\textbf{(1) Chosen-sample Construction}: 
The provided reference answers are utilized as the chosen samples, representing the gold-standard outputs that fully satisfy all specified constraints.

\textbf{(2) Rejected-sample Construction}: 
Rejected samples are those where the model generates results that satisfy the initial and final constraints but contain errors in the intermediate steps. The output length of these samples is controlled to be consistent with the length of the selected samples.

\paragraph{Benchmark Construction Based on LongGenBench and LongEval}
The construction of the benchmark based on LongGenBench and LongEval involves two principal components: the generation schemes used to produce long-form outputs and the construction of chosen and rejected samples.

\textbf{(1) Sequential and Parallel generation schemes}: 
In the sequential mode, each input instruction is decomposed into an ordered sequence of step-wise constraints, and the model incrementally generates intermediate outputs conditioned on the current constraint and accumulated memory. As shown in Figure~\ref{appdix:longeval_case} and Figure~\ref{appdix:longgenbench_case}, in the sequential generation scheme, both chosen and rejected samples indicate that each paragraph should be generated conditioned on the previously generated text. After processing all constraints, the memory states are concatenated to produce the final response.

In the parallel generation setting, the original long instruction is first decomposed into multiple sub-instruction segments. Each segment is then processed in parallel and independently, yielding its corresponding generated output and memory state. Finally, the memory states and outputs produced by all sub-generation processes are aggregated to form a complete generation trajectory. 

\textbf{(2) Chosen and Rejected Sample Construction}: 
To select Chosen samples, the long-form outputs are evaluated in a block-wise manner using a sufficiently capable model, with each segment checked against all input constraints. Only outputs that fully satisfy every specified constraint are designated as correct samples. As illustrated in Figure~\ref{appdix:longeval_case} and Figure~\ref{appdix:longgenbench_case}, the bolded segments of the chosen samples align with the bolded elements in the question.
For the rejected sample, LongEval induces memory errors by perturbing step-wise constraints, whereas LongGenBench simulates memory loss by dropping a subset of constraints. In Figure~\ref{appdix:longeval_case}, the bolded content of the rejected sample fails to match the bolded elements in the question, and in Figure~\ref{appdix:longgenbench_case}, the bolded portions of the question are omitted from the rejected sample. In both cases, an auxiliary model is used to modify the original constraints, leading to hallucinated or missing intermediate memories.

\section{Evaluation Settings}
\label{appdix:evaluation_settings}

\subsection{Prompts for each Task}
As shown in Figure~\ref{appdix:evaluation_prompt}, we provide the prompt used for evaluation. We use different system prompts to adapt the task format. For Long-context Reasoning and Multi-turn Dialogue Understanding, we use the ``System Prompt of Understanding Tasks''. For Long-form Generation, we use the ``System Prompt of Generation Tasks''.
When constructing the evaluation samples, we first implement the user template: The chosen and rejected trajectories in the evaluation preference pair will be randomly shuffled and put into \{Response A\} and \{Response B\}, respectively. Then, we concatenate the system prompt and the implemented user template as the final prompt for every sample. 
 
\begin{figure*}[t]
\begin{AcademicBox}[Evaluation Prompt]
\footnotesize
\textbf{\textit{[System Prompt of Understanding Tasks]}} \\
You are given two memory update trajectories from two different memory-augmented agents (Agent A and Agent B). Each trajectory represents the step-by-step evolution of the agent’s memory as it processes a shared input context and iteratively refines its memory to answer a specific question. Both agents start with an initial memory state and update it over several steps, ultimately producing a final answer. \\
Your task is to compare and evaluate the two memory update paths and select the better agent based on a detailed analysis of their workflows and outputs. Focus on the following aspects:\\
\textbf{Evaluation Criteria:}\\
\textbf{1. Correctness of Each Step in the Work Process:} \\
   - Strict Constraint Adherence: Did the agent follow every specific instruction, numerical constraint, and formatting rule mentioned in the question and context? Failure to meet a hard constraint is a critical failure.\\
   - Check for proper preservation of critical information: Did the agent retain all essential details from the context and question at each step? Identify any step where crucial information was incorrectly added, modified, or discarded. \\
   - Ensure faithfulness: Did any step involve hallucination or fabrication of facts not present in the original context? Each update must strictly derive from the provided input.\\
\textbf{2. Correctness of Each Step vs. Constraints:} \\
   - Constraint Audit: For every step, explicitly check if the agent respected the specific limits. \\
   - Information Fidelity: Identify if any crucial detail was dropped or altered. \\
\textbf{Analysis Instructions:} \\
\textbf{- Overall Judgment:}: Decide which agent has a higher-quality memory update path. Justify your choice by highlighting strengths and weaknesses in their workflows. \\
\textbf{- Explanation:} \\
Analyze both trajectories sequentially. Identify specific instances where one agent demonstrates superior reasoning, information fidelity, or efficiency. For example, note steps where Agent A correctly integrates key details while Agent B omits them, or where Agent B avoids redundancy that Agent A introduces.

- Remember, the goal is to determine which agent is better overall, not necessarily which one is flawless. Even small advantages in step correctness or answer quality can sway the verdict. 

First, output your verdict by strictly following this format: "[[A]]" if Agent A is better, "[[B]]" if Agent B is better. Then provide your explanation.

\hrule \vspace{4pt}

\textbf{\textit{[System Prompt of Generation Tasks]}} \\
You are given two memory update trajectories from two different memory-augmented agents (Agent A and Agent B). Each trajectory represents the step-by-step evolution of the agent’s memory as it processes a shared input context and iteratively refines its memory to answer a specific question. Both agents start with an initial memory state and update it over several steps, ultimately producing a final answer. \\
Your task is to compare and evaluate the two memory update paths and select the better agent based on a detailed analysis of their workflows and outputs. Focus on the following aspects: \\
\textbf{Evaluation Criteria:} \\
\textbf{1. Correctness of Each Step in the Work Process:} \\
   - Strict Constraint Adherence: Did the agent follow every specific instruction, numerical constraint, and formatting rule mentioned in the question and context? Failure to meet a hard constraint is a critical failure.\\
   - Check for proper preservation of critical information: Did the agent retain all essential details from the context and question at each step? Identify any step where crucial information was incorrectly added, modified, or discarded.\\
   - Ensure faithfulness: Did any step involve hallucination or fabrication of facts not present in the original context? Each update must strictly derive from the provided input.\\
\textbf{2. Correctness of Each Step vs. Constraints:} \\
   - Constraint Audit: For every step, explicitly check if the agent respected the specific limits.
   - Information Fidelity: Identify if any crucial detail was dropped or altered. \\
\textbf{Analysis Instructions:} \\
\textbf{- Overall Judgment:} Decide which agent has a higher-quality memory update path. Justify your choice by highlighting strengths and weaknesses in their workflows. \\
\textbf{- Constraint-Based Comparison:} In your step-by-step analysis, use a "Pass/Fail" mindset for the constraints. For example: "Agent A followed the 5-week interval constraint for VR sessions, whereas Agent B missed the interval in Week 30." \\
\textbf{- Explanation:} Analyze both trajectories sequentially. Identify specific instances where one agent demonstrates superior reasoning, information fidelity, or efficiency. For example, note steps where Agent A correctly integrates key details while Agent B omits them, or where Agent B avoids redundancy that Agent A introduces.

- Remember, the goal is to determine which agent is better overall, not necessarily which one is flawless. Even small advantages in step correctness or answer quality can sway the verdict. 

First, output your verdict by strictly following this format: "[[A]]" if Agent A is better, "[[B]]" if Agent B is better. Then provide your explanation.

\hrule \vspace{4pt}

\textbf{\textit{[User Template]}} \\
\lbrack Question\rbrack\\
\{question (along with a long context)\} \\
\lbrack The Start of Agent A's Trajectory\rbrack \{Response A\} \lbrack The End of Agent A's Trajectory\rbrack \\
\lbrack The Start of Agent B's Trajectory\rbrack \{Resposne B\} \lbrack The End of Agent B's Trajectory\rbrack \\
\vspace{-5pt} \hrule \vspace{4pt}
\end{AcademicBox}

\caption{Evaluation Prompt. }
\label{appdix:evaluation_prompt}
\end{figure*}

\subsection{Evaluation Framework}
For proprietary models, we conduct evaluations via the official APIs. 
For open-source models, all evaluations are performed within the LOOM-Scope~\citep{tang2025loom} framework. 
Across all models, we apply the same sampling hyperparameters: we set the top-$p$ value to 0.95, the temperature to 0.7, and the maximum number of generation tokens to 16,384, ensuring that model outputs are not prematurely truncated.
More implementation details are provided in our anonymous code at \url{https://anonymous.4open.science/r/MemRewardBench}.

\section{Details of Ablation Study}

\subsection{Memory Management Patterns}
\label{appdix:memory_management_patterns}

\begin{table}[t]
\centering
\resizebox{0.5\textwidth}{!}{
    \begin{tabular}{l | c c c  | c c c }
        \toprule
        \multirow{2.5}{*}{\textbf{Models}} & \multicolumn{3}{c|}{\textbf{\emph{Sequential}}} & \multicolumn{3}{c}{\textbf{\emph{Parallel}}}  \\
        
        \cmidrule(lr){2-4} \cmidrule(lr){5-7}  

         & LR & LG & \textbf{Avg.} & LR & LG & \textbf{Avg.} \\
        \arrayrulecolor{black}\midrule
        
        GLM4.5-106A22B & 54.9 & 79.7  & 70.8 & 49.5 & 77.6 & 67.7 \\
        Llama-3.3-70B-Instruct & 40.8 & 60.6 & 53.5 & 33.7 & 64.1 & 53.4 \\
        
        Llama-3.1-8B-Instruct & 37.6 & 46.2 & 43.2 & 42.9 & 43.1 & 43.0 \\
        Qwen3-4B & 53.3 & 56.2 & 55.2 & 46.2 & 57.1 & 53.3 \\
        
        \arrayrulecolor{black}\bottomrule
    \end{tabular}
}
\caption{Performance comparison between Sequential and Parallel pattern. "LR", "LG" refer to \textbf{Long-context Reasoning} and \textbf{Long-form Generation} respectively.}
\label{appdix:tab:sp_comparison}
\end{table}

We report a more thorough version of Figure~\ref{fig:sp_comparison} in Table~\ref{appdix:tab:sp_comparison}, which shows that the Long-context Reasoning~(LR) task accuracy of Llama-3.1-8B-Instruct outperforms the Long-form Generation~(LG) task, while the LG task accuracy of Llama-3.3-70B-Instruct outperforms the LR task. 
Given the similarity in patterns between these two tasks, we compute a weighted average of their results to remove task-specific bias, and ultimately conclude that the sequential pattern is easier than the parallel pattern.

\subsection{Failure Case Analysis of Large-scale LLMs}
\label{appdix:abnormal_case}

As shown in Fig~\ref{fig:qwen3_reasoning_case}, the superior performance of Qwen3-14B over both Qwen2.5-72B-Instruct and Llama-3.3-70B-Instruct can be attributed to its significant improvement in reasoning capabilities following post-training. Qwen3-14B excels not only in accurately identifying constraint violations but also in maintaining strict adherence to specified requirements. As demonstrated in the comparison between Agent A and Agent B, Qwen3-14B strikes a careful balance between descriptive quality and instruction adherence, ensuring that each output is both detailed and fully compliant with the established constraints. In contrast, while Llama-3.3-70B-Instruct prioritizes descriptive richness, it struggles with consistently following instructions, leading to inaccurate or poorly structured outputs, as seen in its failure to correctly allocate coffee shops on the required floors. Furthermore, Qwen2.5-72B-Instruct lacks the content precision and accuracy of Qwen3-14B, as evidenced by its less detailed descriptions and the omission of crucial design features.

\begin{figure*}[t] 
\begin{AcademicBox}[Qwen3-14B Performance Case: Enhanced Reasoning Process]
\footnotesize  

\textbf{\textit{Question:}} \\
Construct a skyscraper with 100 floors. Please follow the detailed floor assignments below:

1) Specific floors requirements: - Designate Floor 41 for spa and wellness center use.  - Designate Floor 38 for martial arts dojo use.- Designate Floor 75 for photography studio use. - Designate Floor 21 for daycare center use.  - Designate Floor 35 for private investigator office use.

2) Range floors requirements: Allocate Floors 7 to 11 for a advanced technology and research institute.

3) Periodic floors requirements: Include a coffee shop every 15 floors, starting from Floor 35

4) Document each floor independently with detailed descriptions of the intended facilities, architectural features, and unique design elements.

5) Use '\#*\#' to separate the documentation for each floor (e.g. \#*\# Floor 1: [...] )

Ensure that the document consists of 100 entries, each containing at least 150 words. Ensure clarity and continuity without interruptions or omissions in the narrative throughout the document. When the design of all 100 floors is complete, use '*** finished' to indicate the end of the document. *** started ***\#*\# Floor 1: \\

\textbf{\textit{Respond\_A: [...]}} \\

\textbf{\textit{Respond\_B: [...]}} \\

\vspace{-1ex}
\noindent\rule{\linewidth}{0.3pt}
\vspace{-1ex}
\textbf{\textit{Qwen3-14B:}} \\

Step-by-Step Analysis and Comparison:

1. Specific Floor Requirements: [...]

2. Range Requirements (Floors 7-11): [...]

3. Periodic Requirements (Coffee Shop Every 15 Floors): [...]

4. Document Clarity and Continuity: [...]

5. Information Fidelity and Faithfulness: [...]

6. Overall Document Quality: [...]

Final Verdict:
Agent B produced superior output by strictly following all constraints, ensuring fidelity, and maintaining complete coherent narrative without hallucinations or redundancy. Agent A violated multiple requirements and produced incomplete output. Thus, the better agent is: [[B]].

\vspace{-1ex}
\noindent\rule{\linewidth}{0.3pt}
\vspace{-1ex}
\textbf{\textit{Qwen2.5-72B-Instruct:}} \\

Step-by-Step Analysis:

Floors 1-6 (Initial Floors): [...]

Floors 7-11(Advanced Tech Institute): [...]

Periodic Floors (35-100): [...]

Overall Judgment
Final Verdict: Agent A demonstrates stronger ability to create vivid, immersive narratives despite some redundancy. Better agent: [[A]]

\vspace{-1ex}
\noindent\rule{\linewidth}{0.3pt}
\vspace{-1ex}
\textbf{\textit{Llama-3.3-70B-Instruct:}} \\

The skyscraper design presented by Agent A demonstrates a more comprehensive and detailed approach to meeting the requirements outlined in the prompt. Each floor is meticulously described, with a focus on architectural features, amenities, and the specific needs of various users, from corporate executives and entrepreneurs to wellness enthusiasts and residents. The inclusion of specific details such as materials used for flooring and walls, the type of lighting, and the integration of technology for smart building management showcases a thorough understanding of what makes a modern, functional, and luxurious skyscraper.Agent A's design also prioritizes sustainability, wellness, and community, with the inclusion of green spaces, fitness centers, meditation rooms, and social lounges. This holistic approach to building design not only enhances the quality of life for occupants but also contributes to a healthier and more sustainable urban environment.Furthermore, Agent A's trajectory adheres closely to the specified requirements, including the allocation of specific floors for particular uses (e.g., the spa and wellness center on Floor 41, the photography studio on Floor 75), and incorporates periodic floors for coffee shops starting from Floor 35, as per the instructions. This adherence to the brief demonstrates a strong ability to follow instructions and meet client needs.In contrast, while Agent B's design also showcases luxury and functionality, it lacks the depth and specificity of Agent A's descriptions. Agent B's floors, though well-appointed and modern, sometimes lack the detailed consideration of user experience and the integration of sustainable and wellness-focused elements that are hallmarks of Agent A's design.Therefore, based on the criteria of adherence to instructions, detail and specificity of design, and consideration of user experience and sustainability,  the better agent is: [[A]]

\vspace{-1ex}
\noindent\rule{\linewidth}{0.3pt}
\vspace{-1ex}

\noindent\textbf{\footnotesize Key Insight:} Qwen3-14B demonstrates enhanced post-training reasoning by identifying \textit{constraint violations} that larger models missed.
\end{AcademicBox}
\caption{\footnotesize Case study demonstrating Qwen3-14B's enhanced reasoning process after post-training. The model correctly identifies constraint violations and maintains strict fidelity to requirements, while baseline models prioritize descriptive quality over instruction adherence.}
\label{fig:qwen3_reasoning_case}
\end{figure*}

\subsection{The impact of trajectory length on RM}
\label{appdix:trajectory_length}
The impact of trajectory length manifests in two primary aspects: the accuracy of reward evaluation and the consistency thereof.

Regarding accuracy, we present in Table~\ref{tab:main_result_l} the statistical evaluation results categorized by trajectory length, indicating that model performance invariably declines with longer trajectory lengths.

\begin{table}[t]
\centering
\vspace{-0.5em}

\resizebox{0.5\textwidth}{!}{
\begin{tabular}{l | c c c c c | c }
        \toprule
        \textbf{Models} &  \textbf{8k} & \textbf{16k} & \textbf{32k} & \textbf{64k} & \textbf{128k} & \emph{\textbf{Avg.}}  \\
        \midrule
        \rowcolor{gray!20} \multicolumn{7}{c}{\textit{\faLock~ \textbf{Closed-source Models}}} \\
        \arrayrulecolor{black!20}\midrule
        Claude-Opus-4.5 & 70.9 & \textbf{79.7} & \textbf{78.3} & 73.4 & \textbf{68.8} & \textbf{74.8}  \\ 
        Gemini-3-Pro & \textbf{71.6} & 69.0 & 75.4 & \textbf{75.3} & 63.3 & 71.6  \\ 
        Qwen3-Max & 68.5 & 72.2 & 68.9 & 64.7 & 63.9 & 67.8  \\
        \arrayrulecolor{black}\midrule
        \rowcolor{blue!5} \multicolumn{7}{c}{\textit{\faUnlock~\textbf{Open-source Models}}} \\
        \arrayrulecolor{black!20}\midrule        
        GLM4.5-106A12B & \textbf{67.8} & 63.6 & \textbf{76.7} & \textbf{66.8} & \textbf{63.9} & \textbf{68.2} \\
        Qwen3-235A22B & 64.9 & \textbf{71.8} & \underline{70.1} & \underline{64.8} & 58.5 & \underline{66.6} \\
        Qwen3-32B & 64.0 & 62.0 & 66.3 & 62.0 & \underline{58.7} & 62.9 \\
        Qwen3-14B & 63.0 & 60.4 & 60.8 & 60.4 & 55.9 & 60.3 \\
        Qwen3-8B & 59.0 & 58.4 & 59.7 & 55.1 & 53.6 & 57.3 \\
        Llama3.3-70B & \underline{65.4} & 58.1 & 58.0 & 44.6 & 35.8 & 52.9 \\
        Qwen2.5-72B & 55.1 & 58.1 & 53.2 & 46.2 & 45.6 & 51.8 \\
        Llama3.1-8B & 45.1 & 48.1 & 49.8 & 44.2 & 26.4 & 43.9 \\
        Qwen2.5-7B & 28.9 & 28.8 & 44.2 & 42.8 & 45.6 & 38.2 \\
        \arrayrulecolor{black}\midrule
    \end{tabular}
}
\caption{Results on \texttt{MemoryRewardBench-L(Length Perspective)}.
}
\label{tab:main_result_l}
\end{table}

\begin{figure*}[t]
  \centering
     \includegraphics[width=\linewidth]{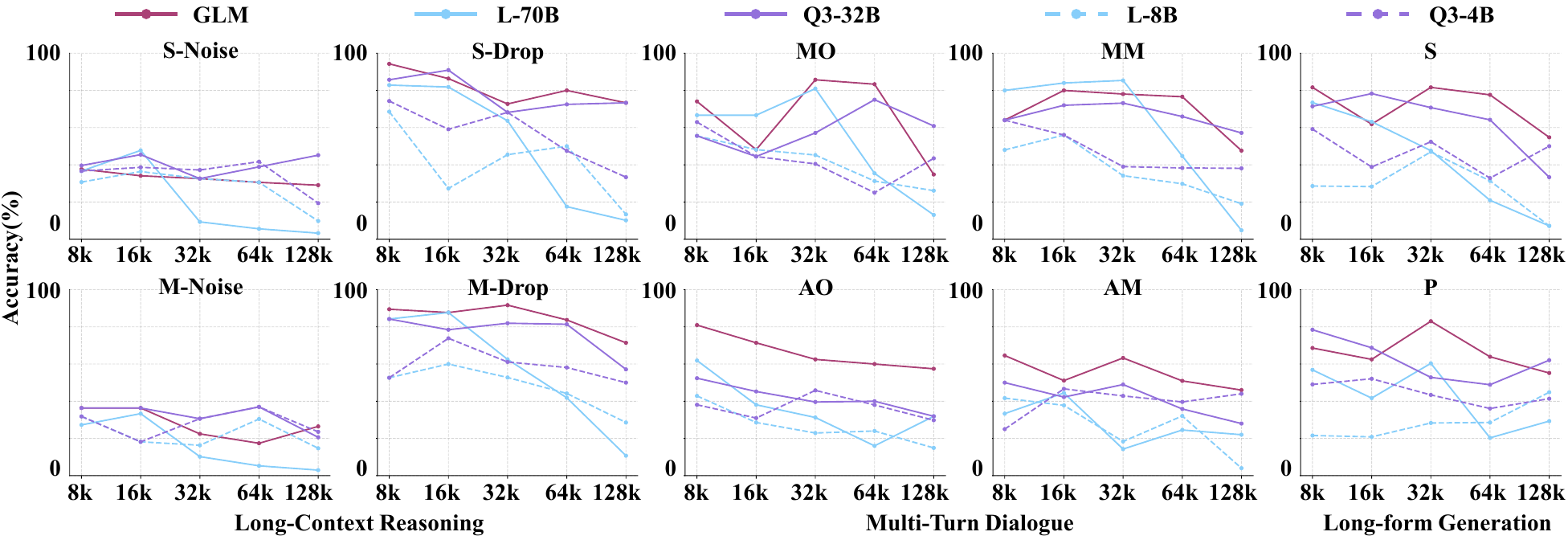}
  \caption{Trends in RM performance and consistency of each subtask with respect to memory management trajectory length.}
\label{fig:length_bias}
\end{figure*}

Regarding consistency, as indicated in Table~\ref{appdix:tab:position_bias}, we initially reversed the contextual order of the chosen and rejected responses during evaluation and tested several models, revealing inconsistencies in their outputs. To investigate this phenomenon, we further conducted consistency experiments across all tasks. As demonstrated in Section~\ref{subsec:effect_memory_management_trajectory}, we provide a more thorough version of length induced bias shown in Figure~\ref{fig:length_bias}. We also report the consistency score of some open-sourced models. As shown in Table~\ref{appdix:tab:main_result_consistency}, models with enhanced reward evaluation capability also tend to exhibit stronger consistency.

\begin{table*}[t]
\centering
\resizebox{\textwidth}{!}{
    \begin{tabular}{l | c c c c c | c c c c c | c c c | c }
        \toprule
    \multirow{2.5}{*}{\textbf{Models}} & \multicolumn{5}{c|}{\textbf{\emph{Long-context Understanding}}} & \multicolumn{5}{c|}{\textbf{\emph{Multi-turn Dialogue}}}  & \multicolumn{3}{c|}{\textbf{\emph{Long-form Generation}}} & \multirow{2.5 }{*}{\emph{\textbf{Avg.}}}  \\
        \cmidrule(lr){2-6} \cmidrule(lr){7-11} \cmidrule(lr){12-14} 
         & S-Noise & S-Drop & M-Noise & M-Drop & avg. & MO & MM & AO & AM & avg. & S & P & avg. &\\
         \midrule

        GLM4.5-106A12B & 34.12 & \textbf{82.84} & 26.09 & \textbf{86.34} & \textbf{55.25} & \textbf{69.46} & \textbf{71.7} & \textbf{55.1} & \textbf{65.94} & \textbf{64.5} & \textbf{71.99} & \textbf{65.89} & \textbf{69.38} & \textbf{63.04} \\
        Qwen2.5-72B & 21.57 & \underline{74.63} & 22.28 & \underline{74.45} & 45.63 & 46.71 & 50.31 & 25.71 & \underline{37.99} & 38.50 & 44.20 & 39.94 & 42.38 & 42.17 \\
        Llama-3.3-70B & 24.71 & 50.00 & 9.78 & 60.79 & 35.75 & 53.29 & \underline{59.12} & 26.94 & 31.88 & \underline{40.25} & 47.05 & 34.40 & 41.63 & 39.21 \\
        Qwen3-8B & \textbf{40.78} & 47.76 & \textbf{41.85} & 61.23 & 48.00 & \underline{55.09} & 52.83 & 27.35 & 27.51 & 38.25 & \underline{59.74} & 51.02 & 56.00 & 47.42 \\
        Llama-3.1-8B & 27.06 & 44.03 & 19.02 & 50.22 & 34.63 & 38.92 & 33.33 & 24.08 & 22.27 & 28.50 & 27.13 & 20.70 & 24.38 & 29.17 \\
        Qwen2.5-7B & 10.20 & 15.67 & 20.11 & 22.91 & 17.00 & 31.74 & 34.59 & 13.88 & 14.41 & 21.88 & 32.82 & 31.20 & 32.13 & 23.67 \\
        Qwen3-4B & 35.69 & 58.96 & 28.80 & 62.56 & 45.63 & 39.52 & 44.03 & \underline{37.55} & 36.25 & 38.88 & 44.86 & 41.98 & 43.63 & 42.71 \\

        \arrayrulecolor{black}\midrule
    \end{tabular}
}
\caption{Results on \texttt{MemoryRewardBench-Consistency}. "S."  and  "P." refer to "Sequential" and "Parallel" respectively. "SC." and "MI." refer to "Self-Correct" and "Mask-Info" respectively.
}
\label{appdix:tab:main_result_consistency}
\end{table*}

\subsection{Global Constraint of Long-form Generation}
\label{appdix:long_form_gen_detail_ablation}

\begin{figure*}[t]
  \centering
  \begin{subfigure}[t]{0.32\textwidth}
    \centering
     \includegraphics[width=\linewidth]{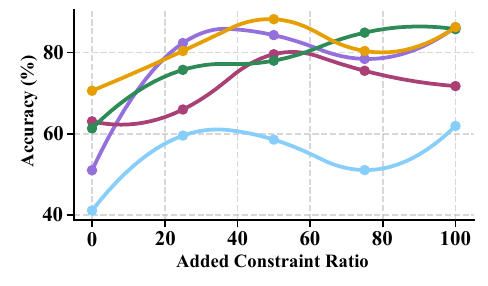}
    \caption{LongEval}
    \label{fig:position_bias_self_correct}
  \end{subfigure}
  \hfill
  \begin{subfigure}[t]{0.32\textwidth}
    \centering
    \includegraphics[width=\linewidth]{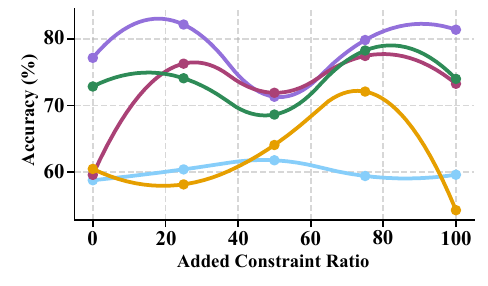}
    \caption{LongGenBench}
    \label{fig:position_bias_mask_info}
  \end{subfigure}
  \hfill
  \begin{subfigure}[t]{0.32\textwidth}
    \centering
    \includegraphics[width=\linewidth]{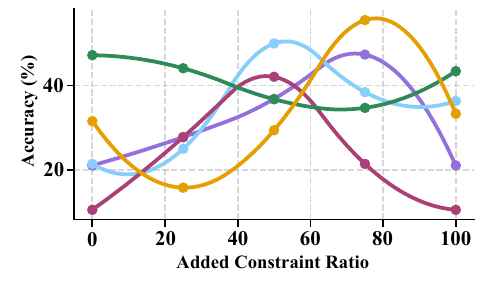}
    \caption{LongProc}
    \label{fig:position_bias_mask_info}
  \end{subfigure}
  \vspace{-0.5em}
  \caption{Overall performance trends on LongEval, LongGenBench, and LongProc as the constraint density varies in long-form generation tasks.} 
  \label{fig:longgen_ablation_compare}
\end{figure*}

As shown in Figure~\ref{fig:longgen_ablation_compare}, we present overall performance trends of LongEval, LongGenBench, and LongProc on long-form generation tasks as the constraint density varies. A unified constraint density control strategy is adopted across all three datasets to ensure comparability across task settings.
For the \textbf{LongProc} dataset, we vary the availability of route constraints by progressively partitioning the complete set of known routes. 
A constraint rate of 0\% corresponds to the setting where no known routes are provided to the model, in which case the RMs infer the path solely from the given start and end points. 
In contrast, a constraint rate of 100\% supplies the model with the full set of known routes, enabling the reward model to be fully informed of all possible path transitions.

For \textbf{LongEval} and \textbf{LongGenBench}, constraint density is controlled by selecting subsets of the complete constraint set while preserving the semantic integrity of each individual constraint. 
As shown in Figure~\ref{appdix:longgen_ablation_case}, under the 0\% constraint setting, the model is provided only with a minimal prompt to generate long-form text, without any explicit content or structural requirements.
At the 100\% constraint level, all predefined constraints are supplied.
Intermediate constraint ratios are obtained by proportionally sampling from the full constraint set, enabling a systematic analysis of RMs behavior under varying degrees of constraint supervision.
\begin{figure*}[t]
\begin{AcademicBox}[LongGen Ablation]
\footnotesize
\textbf{\textit{[0\%]}} \\
Generate a complete weekly diary for Benjamin for the entire year of 2018\\
\textbf{\textit{[25\%]}} \\
Benjamin is a architect with a passion for chronicling his vibrant life through weekly diary entries. Captures:1) Family member birthday: husband (yourself) (birthday on December 07), my wife (birthday on November 21)\\
\textbf{\textit{[50\%]}} \\
Benjamin is a architect with a passion for chronicling his vibrant life through weekly diary entries. Captures:1) Family member birthday: husband (yourself) (birthday on December 07), my wife (birthday on November 21), my child\_1 (birthday on September 21), my child\_2 (birthday on January 19), my child\_3 (birthday on November 20)\\
\textbf{\textit{[75\%]}} \\
Benjamin is a architect with a passion for chronicling his vibrant life through weekly diary entries. Captures:1) Family member birthday: husband (yourself) (birthday on December 07), my wife (birthday on November 21), my child\_1 (birthday on September 21), my child\_2 (birthday on January 19), my child\_3 (birthday on November 20), 2) Attending a traditional dance festival in India in week 30-31.3) Attend a DIY pottery and ceramics class every 5 weeks on weekends, starting from week 11.\\
\textbf{\textit{[100\%]}} \\
Benjamin is a architect with a passion for chronicling his vibrant life through weekly diary entries. Captures:1) Family member birthday: husband (yourself) (birthday on December 07), my wife (birthday on November 21), my child\_1 (birthday on September 21), my child\_2 (birthday on January 19), my child\_3 (birthday on November 20), 2) Attending a traditional dance festival in India in week 30-31.3) Attend a DIY pottery and ceramics class every 5 weeks on weekends, starting from week 11. 4) Weekly updates on weather changes, work developments, family life, and other interesting topics.5) Use '\#\*\#' to separate each weekly entry (e.g. \#\*\# Week 1 (January 1st - January 7th): \\

\vspace{-5pt} \hrule \vspace{4pt}
\end{AcademicBox}

\caption{LongGen Ablation. }
\label{appdix:longgen_ablation_case}
\end{figure*}

\subsection{Multi-turn Understanding with Auxiliary Signals}
\label{appdix:multiturn_dialogue_under_tag}
We investigate the role of auxiliary signals, such as tags, in the A-Mem memory system~\citep{xu2025mem}. 
As illustrated in Figure~\ref{fig:rm_comparison_auxiliary_signals}, we present two scenarios: one where auxiliary signals are used, and another where they are absent. In the first scenario, A-Mem utilizes semantic tags to summarize dialogue content; in the second, these tags are removed, causing them to lose part of their semantic function. This comparison highlights the impact of auxiliary signals on the system's performance.

The experimental results show that when the data contains meaningful semantic tags, the model is able to more effectively distinguish between different memory processes and dialogue segments, leading to significant improvements in performance on the memsys task. As shown in Figure~\ref{appndix:tab:tags_impact}, tags in the A-Mem system not only serve as metadata for structured memory but also play a critical role in semantic indexing and organization. They provide additional context for the model, helping it locate key information and understand the structure of long conversations, thereby optimizing memory management.

\begin{table}[t]
  \centering 
  \resizebox{0.9\linewidth}{!}{
    
  \begin{tabular}{l c c c}
    \toprule 
    \multirow{2}{*}{Models} & \multicolumn{2}{c}{subtasks} & \multirow{2}{*}{avg} \\
    \cmidrule(lr){2-3}    
                            & With & WithOut &     \\
    \midrule 
        GLM-4.5-Air             & \textbf{0.759} & \textbf{0.620} & \textbf{0.690} \\
        Qwen3-14B   & 0.690 & 0.540 & 0.603 \\
        Qwen2.5-14B-Instruct   & 0.621 & 0.500 & 0.567 \\
        Qwen3-4B                & 0.655 & 0.510 & 0.585 \\

    \bottomrule  
  \end{tabular}
  }
  \caption{Impact of structured tags on LLM performance in memsys subtasks 
           (WithTags: memsys\_v2 with structured tags; WithOutTags: memsys without tags)}
  \label{appndix:tab:tags_impact}
\end{table}

In contrast, removing the tags significantly impairs the model's ability to distinguish memory processes, resulting in a decline in performance. This validates the importance of designing effective memory organization mechanisms (such as tagging) when building memory systems to enhance a model's ability to comprehend long conversations.

\begin{figure}[t]  
    \centering
    \begin{minipage}{0.48\textwidth}  
        \centering
        \begin{AcademicBox}[Tags Comparison Case]
        \footnotesize

        \textbf{Case Overview:}

        \textbf{Case 1: RM Evaluation with Auxiliary Signals}  

        \textbf{Conversation:}

          \textbf{Context:} The speaker offers emotional support, expressing a willingness to listen to a friend or acquaintance in a caring and empathetic manner.

          \textbf{Keywords:} ['support', 'conversation', 'emotional']

          \textbf{Tags:} ['emotional-support', 'personal-communication', 'friendship']

        \textbf{Case 2: RM Evaluation without Auxiliary Signals} 

        \textbf{Conversation:}

          \textbf{Context:} The speaker offers emotional support, expressing a willingness to listen to a friend or acquaintance in a caring and empathetic manner.

          \textbf{Keywords:} ['element1', 'element2', 'element3']

        \end{AcademicBox}
    \end{minipage}
    \caption{Comparison of RM evaluation for multi-turn dialogue understanding in a memory system, with and without auxiliary signals.}
    \label{fig:rm_comparison_auxiliary_signals}
\end{figure}


\begin{table}[t]
\centering
\resizebox{0.5\textwidth}{!}{
    \begin{tabular}{l | c c c  | c c c }
        \toprule
        \multirow{2.5}{*}{\textbf{Models}} & \multicolumn{3}{c|}{\textbf{\emph{Chosen-First}}} & \multicolumn{3}{c}{\textbf{\emph{Rejected-First}}}  \\
        
        \cmidrule(lr){2-4} \cmidrule(lr){5-7}  

         & LU & MD & \textbf{Avg.} & LU & MD & \textbf{Avg.} \\
        \arrayrulecolor{black}\midrule
        
        GLM-4.5-Air & 84.1 & 52.4 & 68.3 & 53.9 & 63.4 & 58.6 \\ 
        Llama-3.3-70B-Ins & 65.6 & 58.6 & 62.1 & 39.6 & 27.5 & 33.6 \\
        Llama-3.1-8B-Ins & 56.8 & 38.5 & 47.6 & 40.1 & 31.4 & 35.8 \\
        Qwen3-4B & 61.5 & 37.2 & 49.4 & 57.9 & 52.1 & 55.0 \\
        
        \arrayrulecolor{black}\bottomrule
    \end{tabular}
 }
\caption{Performance comparison on MultiHop-QA task. }
\label{appdix:tab:position_bias}
\end{table}

        

        
        

\end{document}